\documentclass[10pt,twocolumn,letterpaper]{article}

\usepackage{iccv}
\usepackage{times}
\usepackage{epsfig}
\usepackage{graphicx}
\usepackage{amsmath}
\usepackage{amssymb}

\usepackage{color}
\usepackage{latex_pkgs/maths}
\usepackage{latex_pkgs/msformatting}
\usepackage{algorithm2e}
\usepackage{listings}
\usepackage{cuted} 

\renewcommand{\vec}[1]{\mathbf{#1}} 
\newcommand{\tb}[1]{\textbf{#1}}

\newcommand{\SU}[1]{\textcolor{red}{#1}}

\usepackage[pagebackref=true,breaklinks=true,letterpaper=true,colorlinks,bookmarks=false]{hyperref}

\iccvfinalcopy 

\def\iccvPaperID{1935} 

\begin{document}
\title{AMTnet: Action-Micro-Tube Regression by\\End-to-end Trainable Deep Architecture}

\author{Suman Saha \qquad Gurkirt Singh \qquad Fabio Cuzzolin \\ 
Oxford Brookes University, Oxford, United Kingdom\\ 
{\tt\small \{suman.saha-2014, gurkirt.singh-2015, fabio.cuzzolin\}@brookes.ac.uk}
}
\maketitle
\begin{abstract}

Dominant approaches to action detection can only provide {sub-optimal} solutions to the problem,
as they rely on seeking frame-level detections, to later compose them into `action tubes' in a post-processing step. With this paper we radically depart from current practice, and take a first step towards the design and implementation of a deep network architecture able to classify and regress whole video subsets, so providing a truly optimal solution of the action detection problem.
In this work, in particular, we propose a novel deep net framework able to regress and classify 3D region proposals spanning two {successive} video frames, whose core is an evolution of classical region proposal networks (RPNs). As such, our 3D-RPN net is able to effectively encode the temporal aspect of actions by purely exploiting appearance, as opposed to methods which heavily rely on expensive flow maps. 
The proposed model is end-to-end trainable and can be jointly optimised for
action localisation and classification in a single step. 
At test time the network predicts `micro-tubes' encompassing two {successive} frames, which are linked up into complete action tubes via a new algorithm 
which exploits the temporal encoding learned by the network and cuts computation time by 50\%.
Promising results on the J-HMDB-21 and UCF-101 action detection datasets show that
our model does outperform the state-of-the-art when relying purely on appearance.

\vspace{-3mm}

\end{abstract}
\section{Introduction} \label{sec:intro}
In recent years 
most action detection frameworks
~\cite{Georgia-2015a,Weinzaepfel-2015,peng2016eccv,Saha2016} employ deep convolutional neural network (CNN) architectures,
mainly based on 
region proposal algorithms~\cite{uijlings-2013,zitnick2014edge,ren2015faster} and two-stream RGB and optical flow CNNs~\cite{Simonyan-2014,Georgia-2015a}.
These methods first construct training hypotheses by generating region proposals (or `regions of interest', ROI\footnote{A ROI is a rectangular bounding box parameterized as 4 coordinates in a 2D plane $[x1$ $y1$ $x2$ $y2]$.}), using either Selective Search~\cite{uijlings-2013}, EdgeBoxes~\cite{zitnick2014edge} or a region proposal network (RPN)~\cite{ren2015faster}.
ROIs are then sampled as positive and negative training examples as per the ground-truth.
Subsequently, CNN features are extracted from each region proposal. Finally,
ROI pooled features are fed to a softmax and a regression layer for action classification and bounding box regression, respectively.

This dominant paradigm for action detection~\cite{Georgia-2015a,Weinzaepfel-2015,peng2016eccv,Saha2016}, however,
only provides a \emph{sub-optimal} solution to the problem. 
Indeed, rather than solving for 
\begin{equation} \label{eq:action-detection-problem}
T^* \doteq \arg \max_{T \subset V} \text{score}(T), 
\end{equation}
where $T$ is a subset of the input video of duration $D$ associated with an instance of a known action class, 
they seek partial solutions for each video frame
$
R^*(t) \doteq \arg \max_{R \subset I(t)} \text{score}(R),
$
to later compose in a post-processing step partial frame-level solutions into a solution $\hat{T} = [R^*(1),...,R^*(D)]$ of the original problem (\ref{eq:action-detection-problem}), typically called \emph{action tubes}~\cite{Georgia-2015a}. By definition, score$(\hat{T})\leq$score$(T^*)$ and such methods are bound to provide suboptimal solutions.
The post-processing step is essential as those CNNs do not learn the temporal
associations between region proposals belonging to successive video frames.
This way of training is mostly suitable for object detection, 
but inadequate for action detection where both spatial and temporal localisation are crucial.
To compensate for this and learn the temporal dynamics of human actions,
optical flow features are heavily exploited~\cite{Georgia-2015a,Weinzaepfel-2015,peng2016eccv,Saha2016}.


\begin{figure*}[t]
  \centering
  \includegraphics[scale=0.34]{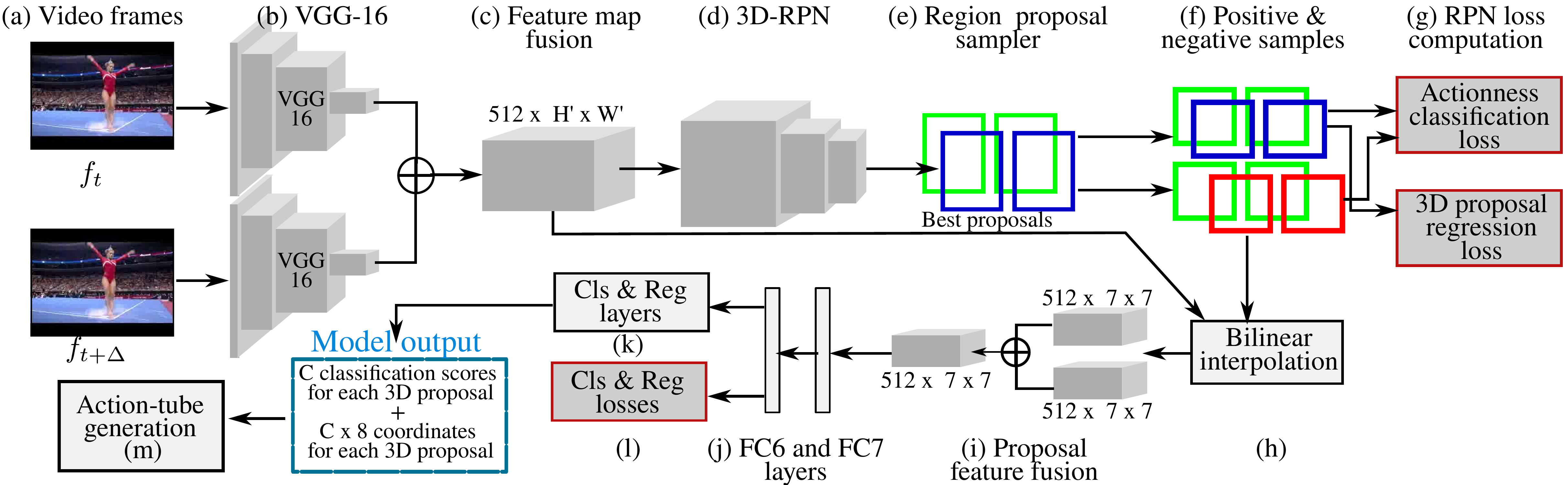}
  \vskip 2mm
  \caption{
  {\small
      \textit{
  At train time, the input to the network is a pair of successive video frames~\textbf{(a)} which are processed through two parallel VGG-16 networks~\textbf{(b)}.
  The feature maps generated by the last convolution layers are fused~\textbf{(c)} and the fused feature map is fed to a 3D-RPN network~\textbf{(d)}.
  The RPN generates 3D region proposals and their associated \emph{actionness}~\cite{chen2014actionness} scores
  which are then sampled as positive and negative training examples~\textbf{(f)} by a proposal sampler~\textbf{(e)}.
  The sampled proposals and their scores are used to compute the \emph{actionness} and 3D proposal regression losses~\textbf{(g)}.
  Subsequently, a bilinear feature pooling~\textbf{(h)} and an element-wise feature fusion~\textbf{(i)} are used to obtain 
  a fixed sized feature representation for each sampled 3D proposal.
  Finally, the pooled and fused features are passed through fully connected (FC6 \& FC7)~\textbf{(j)},
  classification and regression~\textbf{(k)} layers to train for action classification and a micro-tube regression.  
  {At test time, the predicted micro-tubes are linked in time by the action-tube generator~\textbf{(m)}}.
  }
  }
  }
  \label{fig:algorithmOverview} \vspace{-4mm}
\end{figure*}

\begin{figure}[h!]
  \centering
  \includegraphics[scale=0.4]{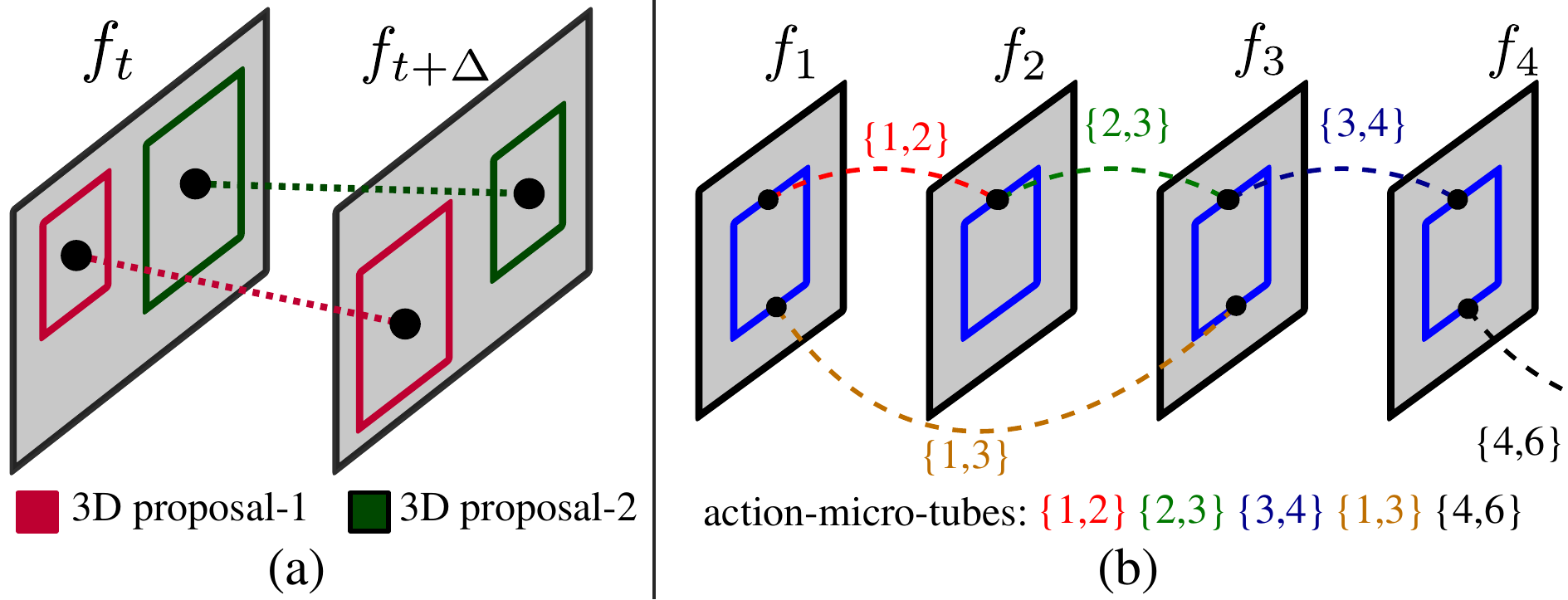}
  \vskip 2mm
  \caption{
  {\small
      \textit{
  \textbf{(a)} The 3D region proposals generated by our 3D-RPN network span pairs of successive video frames $f_{t}$ and $f_{t+\Delta}$ at temporal distance $\Delta$.
  \textbf{(b)} Ground-truth action-micro-tubes generated from different pairs of successive video frames.}
  }
  }
  
  \label{fig:3d-prop} \vspace{-4mm}
\end{figure}

With this paper we intend to initiate a research programme leading, in the medium term, to a new deep network architecture able to classify and regress whole video subsets. In such a network, the concepts of (video) region proposal and action tube will coincide.
\\
In this work, in particular, we take a first step towards a truly optimal solution of the action detection problem by considering video region proposals formed by a pair of bounding boxes spanning two successive video frames at an arbitrary temporal interval $\Delta$ 
(see Figure~\ref{fig:3d-prop}).
We call these pairs of bounding boxes \emph{3D region proposals}. 
The advantages of this approach are that
a) appearance features can be exploited to learn temporal 
dependencies (unlike what happens in current approaches), thus
boosting detection performance;
b) the linking of frame-level detections over time is no longer a 
post processing step and can be (partially) learned by the network.
Obviously, at this stage we still need to construct action tubes from 3D region proposals.

We thus propose a radically new approach to action detection based on (1) a novel deep learning architecture for regressing and classifying two-frame \emph{micro-tubes}\footnote{We call `micro-tubes' the 3D video region proposals, spanning pairs of {successive} frames, generated by the network at test time.}, illustrated in Figure \ref{fig:algorithmOverview}, in combination with (2) an original strategy for linking micro-tubes up into proper action tubes.
At test time, this new framework does not completely rely on post-processing for assembling frame-level detections, but 
makes use of the temporal encoding learned by the network.
\\
We show that:
i) such a network trained on pairs of successive RGB video frames can learn the spatial and temporal extents 
of action instances relatively better than those trained on individual video frames, and 
ii) our model outperforms the current state-of-the-art~\cite{Georgia-2015a,Weinzaepfel-2015,Saha2016}
in spatio-temporal action detection by just exploiting appearance (the RGB video frames), in opposition to the methods
which heavily exploit expensive optical flow maps.

Just to be clear, the aim of this paper is not to renounce to optical flow cues, but to move from frame-level detections to whole tube regression.
Indeed the method can be easily extended to incorporate motion at the micro-tube level rather than frame level, allowing fusion of appearance and motion at training time, unlike current methods \cite{peng2016eccv,Saha2016}.

\textbf{Overview of the approach}. 
Our proposed network architecture (see Figure~\ref{fig:algorithmOverview}) employs and adapts some of the architectural components recently proposed in~\cite{ren2015faster,Johnson2016}.
\\
At training time, the input to the model is a pair of successive video frames~\textbf{(a)} 
which are fed to two parallel CNNs~\textbf{(b)}~(\S~Section~\ref{subsec:vgg16-fm-fusion}).
The output feature maps of the two CNNs are fused~\textbf{(c)} and 
passed as input to a 3D region proposal network (3D-RPN)~\textbf{(d)}~(\S~Section~\ref{subsec:3d-rpn}).
The 3D-RPN network generates 3D region proposals and their associated \emph{actionness}\footnote{The term \emph{actionness}~\cite{chen2014actionness} is used to denote the possibility of an action being present within a 3D region proposal.} \cite{chen2014actionness} scores,
which are then sampled as positive and negative training examples~\textbf{(f)} by a proposal sampler~\textbf{(e)}~(\S~\ref{subsec:prop-sampler}).
A training mini-batch of $256$ examples are constructed from these positive and negative samples.
The mini-batch is firstly used to compute the \emph{actionness} classification and 3D proposal regression losses~\textbf{(g)}~(\S~\ref{subsec:loss-func}),
and secondly, to pool CNN features (for each 3D proposal) using a bilinear interpolation layer~\textbf{(h)}~(\S~\ref{bilinear-interpolation}).

In order to interface with the fully connected layers~\textbf{(j)}~(\S~\ref{subsec:fc-layers}), 
bilinear interpolation is used to get a fixed-size feature representation for each variably sized 3D region proposal.
As our 3D proposals consist of a pair of bounding boxes, 
we apply bilinear feature pooling independently on each bounding box in a pair, which gives rise to two
fixed-size pooled feature maps of size $[512 \times kh \times kw]$, where $kh$ $=$ $kw$ $=$ $7$ for each 3D proposal.
We then apply element-wise fusion~\textbf{(i)}~(\S~\ref{bilinear-interpolation}) to these $2$ feature maps.
Each pooled and then fused feature map (representing a 3D proposal) is passed to two fully connected layers (FC6 and FC7))~\textbf{(j)}~(\S~\ref{subsec:fc-layers}).
The output of the FC7 layer is a fixed sized feature vector of shape $[4096 \times 1]$.
These $4096$ dimension feature vectors are then used by a classification and a regression layers~\textbf{(k)}~(\S~\ref{subsec:fc-layers}) to output
\textbf{(1)} $B \times C$ classification scores and  
\textbf{(2)} $B \times C \times 8$ coordinate values where 
$B$ is the number of 3D proposals in a training mini-batch and $C$ is the number of action categories in a given dataset.
\\
\indent At test time we select top $1000$ predicted micro-tubes by using non-maximum suppression, modified to work with pairs of bounding boxes
and pass these to an action-tube generator~\textbf{(m)}~(\S~\ref{sec:action-tube}) which links those micro-tubes in time.
At both training and test time, our model receives as input successive video frames $f_{t}$, $f_{t+\Delta}$. 
At training time we generate training pairs using 2 different $\Delta$ values $1$ and $2$ (\S~\ref{sec:data_eval_samp}).
At test time we fix $\Delta=1$.
As we show in the Section~\ref{subsec:discussion}, even consecutive frames ($\Delta=1$) carry significantly different information which affects the overall video-mAP. 
Throughout this paper, 
``3D region proposals'' denotes the RPN-generated pairs of bounding boxes 
regressed by the middle layer (Figure~\ref{fig:algorithmOverview}~\textbf{(g)}),
whereas ``micro-tubes'' refers to the 3D proposals regressed by
the end layer (Figure~\ref{fig:algorithmOverview}~\textbf{(l)}).

\textbf{Contributions.}
In summary, the key contributions of this work are:
\textbf{(1)} on the methodological side, a key conceptual step forward from action detection paradigms relying on frame-level region proposals towards networks able to regress optimal solutions to the problem;
\textbf{(2)} a novel, end-to-end trainable deep network architecture which addresses the spatiotemporal action localisation and classification
task jointly using a single round of optimisation;
\textbf{(3)} at the core of this architecture, a new design for a fully convolutional action localisation network (3D-RPN) which generates 3D video region proposals rather than frame-level ones;
\textbf{(4)} a simple but efficient regression technique for regressing such 3D proposals;
\textbf{(5)} a new action-tube generation algorithm suitable for connecting the micro-tubes so generated, which exploits the temporal encoding learnt by the network.
\\
\indent Experimental results on the J-HMDB-21 and UCF-101 action detection datasets show that
our model outperforms state-of-the-art appearance-based models, while being competitive with methods using parallel appearance and flow streams.
Finally, to the best of our knowledge,
this is the first work in action detection which uses \emph{bilinear interpolation}~\cite{gregor2015draw,jaderberg2015spatial}
instead of the widely used RoI max-pooling layer~\cite{girshick2015fast},
thus allowing gradients to flow backwards for both convolution
features and coordinates of bounding boxes.

\section{Related work}

Deep learning architectures have been increasingly applied of late to action classification \cite{Shuiwang-2013,Karpathy-2014,Simonyan-2014,tran2014learning}, 
spatial \cite{Georgia-2015a}, temporal~\cite{Shou2016} and spatio-temporal~\cite{Weinzaepfel-2015,Saha2016,peng2016eccv}
action localisation.
While many works concern either spatial action localisation \cite{Lu_2015_CVPR,wangcvpr2016,jain2014supervoxel,Soomro2015} 
in trimmed videos or temporal localisation~\cite{laptev2007retrieving,gaidon2013temporal,tian2013spatiotemporal,oneata2014efficient,wang2015robust,
Shou2016,yeungcvpr2016,yeung2015every} in untrimmed videos,
only a handful number of methods have been proposed to tackle both problems jointly. 
Spatial action localisation has been mostly addressed using segmentation~\cite{Lu_2015_CVPR,Soomro2015,jain2014supervoxel}
or by linking frame-level region proposal ~\cite{Georgia-2015a,Weinzaepfel-2015,wangcvpr2016}. 
Gkioxari and Malik \cite{Georgia-2015a}, in particular, have built on \cite{girshick-2014} and \cite{Simonyan-2014} 
to tackle spatial action localisation in temporally trimmed videos,
using Selective-Search~\cite{uijlings-2013} based region proposals on each frame of the videos.


Most recently, supervised frame-level action proposal generation and classification have been used by 
Saha \etal~\cite{Saha2016} and Peng \etal~\cite{peng2016eccv}, via a 
Faster R-CNN~\cite{ren2015faster} object detector, to
generate frame level detections independently for each frame and link them in time in a post-processing step. 
Unlike~\cite{vanGemert2015apt,Georgia-2015a,Weinzaepfel-2015}, current methods \cite{wangcvpr2016,Saha2016,peng2016eccv} are able to leverage on end-to-end trainable deep-models~\cite{ren2015faster} for frame level detection.
However, tube construction is still tackled separately from region proposal generation. 

Our novel network architecture, generates micro-tubes (the smallest possible video-level region proposals) which span across frames, and are labelled using a single soft-max score vector, in opposition to~\cite{Georgia-2015a,Weinzaepfel-2015,peng2016eccv,Saha2016} which generate frame-level region proposals.
Unlike~\cite{Georgia-2015a,Weinzaepfel-2015,peng2016eccv,Saha2016},
our proposed model is \emph{end-to-end trainable} and requires a \emph{single step of optimisation} per training iteration.
To the contrary,
\cite{Georgia-2015a,Weinzaepfel-2015} use a multi-stage training strategy mutuated from R-CNN object detection~\cite{girshick-2014} 
which requires training two CNNs (appearance and optical-flow) independently, plus
a battery of SVMs. The two most recent papers
\cite{peng2016eccv,Saha2016} extend this Faster R-CNN~\cite{ren2015faster} framework
and train independently appearance and motion CNNs.
Compared to~\cite{Georgia-2015a,Weinzaepfel-2015,peng2016eccv,Saha2016},
which heavily exploit expensive optical flow maps,
our model {learns spatiotemporal feature encoding directly from raw RGB video frames}.

\section{Network Architecture} \label{sec:model-architecture}
\vskip -2mm
All the stages of 
Figure \ref{fig:algorithmOverview} are described below in detail.
\begin{figure*}[t!]
  \centering
  \includegraphics[scale=0.75]{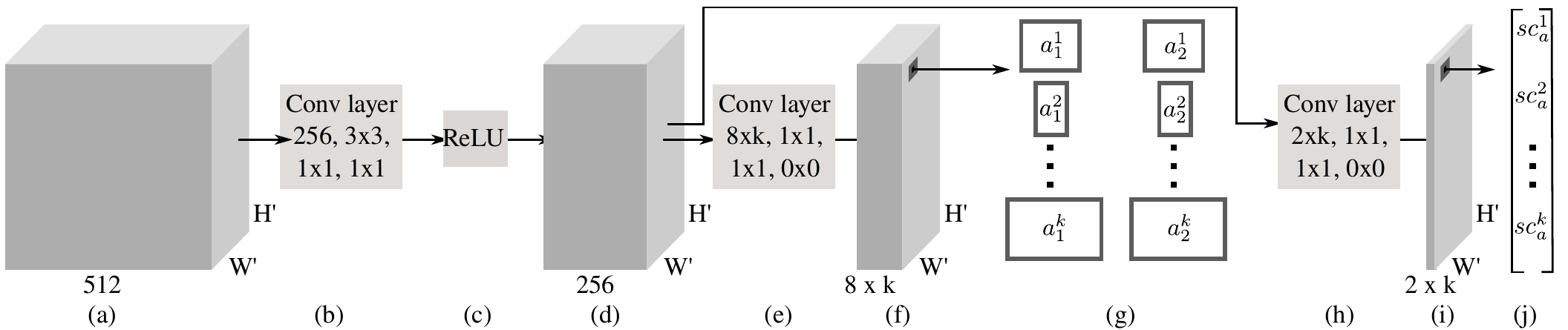}
  \vskip 2mm
  \caption{
   {\small
      \textit{
  3D-RPN  architecture.
  }
  }
  }
  \label{fig:3d-rpn} \vspace{-4mm}
\end{figure*}
\subsection{Convolutional Neural Network} \label{subsec:vgg16-fm-fusion}
The convolutional (conv) layers of our network follow the VGG-16 architecture~\cite{Simonyan-2014}.
We use two parallel VGG-16 networks (\S~Figure~\ref{fig:algorithmOverview}~\textbf{(b)}) to apply 
convolution over a pair of successive video frames.
Each VGG-16 has $13$ conv layers intermixed with $5$ max pooling layers.
Each conv layer has a $3 \times 3$ filter and $1 \times 1$ stride and padding. 
Each max pooling layer has filter shape $2 \times 2$.
We discard all the VGG-16 layers after the last ($13$-th) conv layer.

\textbf{Feature map fusion.}
Our network takes two successive video frames $f_{t}$ and $f_{t+\Delta}$ as inputs.
For a input video frame of shape $[3 \times H \times W]$,
the last conv layer of each VGG-16 outputs a feature map of 
shape $[D \times H' \times W']$ where $D=512$, $H'=\frac{H}{16}$, and  $W'=\frac{W}{16}$.
We fuse the two conv feature maps produced by the two parallel VGG-16 networks
using element-wise sum fusion~(\S~Figure~\ref{fig:algorithmOverview}~\textbf{(c)}).
\\ As a consequence, the fused feature map encodes both appearance \emph{and} 
motion information (for frames $f_{t}$ and $f_{t+\Delta}$), which we pass as input to our 3D-RPN network.

Our new 3D region proposal network~(Figure~\ref{fig:algorithmOverview}~\textbf{(d)}) builds on the basic RPN structure	~\cite{ren2015faster} to propose a fully convolutional network which can generate 3D region proposals via a number of 
significant architectural changes.

\subsection{3D region proposal network} \label{subsec:3d-rpn}
\textbf{3D region proposal generation.} 
As we explained,
unlike a classical RPN~\cite{ren2015faster} which generates region proposals (rectangular bounding boxes) per image,
our 3D-RPN network generates (video) region proposals spanning a pair of video frames.
A single proposal thus consists of a pair of rectangular bounding boxes.
The input to our 3D-RPN 
is a fused VGG-16 feature map~(\S~Figure~\ref{fig:algorithmOverview}~\textbf{(c)}) of size $[512 \times H' \times W']$.
We generate anchor boxes in a similar way as in~\cite{ren2015faster}: namely,
we project back each point in the $H' \times W'$ grid (of the input feature map)
onto the original image plane of size $H \times W$.
For each projected point we generate $k$ pairs of anchor boxes of different aspect ratios.

Let $(x_{a_{i}}$,$y_{a_{i}}$,$w_{a_{i}}$,$h_{a_{i}})$ denote
the centroid, width and height of the anchor boxes in a pair. 
We use the subscript $i$ to index the two boxes in a pair, i.e. $i=\{1,2\}$. 
Similarly, $(x_{g_{i}},y_{g_{i}},w_{g_{i}},h_{g_{i}})$ refer to the
centroid, width and height of the ground-truth pair.
We can transform a pair of input anchor boxes into a predicted pair of ground-truth boxes via\footnote{We removed the subscript $i$ in Eq.~\ref{eq:transform} for sake of simplicity.}:
\begin{align} \label{eq:transform}
& x_g = x_a + \phi_x w_a & y_g = y_a + \phi_y h_a \nonumber \\
& w_g = w_a \exp(\phi_w)  & h_g = h_a \exp(\phi_h) 
\end{align}
where $(\phi_{x_{i}},\phi_{y_{i}})$ specify a scale-invariant translation of the center of the anchor boxes, and
$(\phi_{w_{i}},\phi_{h_{i}})$ specify a log-space translation of their width and height.

Both RPN and the micro-tube regression layer (Figure~\ref{fig:algorithmOverview}~\textbf{(k)}) 
predict the bounding box regression offsets $(\phi_{x_{i}},\phi_{y_{i}},\phi_{w_{i}},\phi_{h_{i}})$.
Our anchor generation approach differs from that of~\cite{ren2015faster}, in the sense that 
we generate $k$ \emph{pairs} of anchors instead of $k$ anchors.

\textbf{Network architecture.} 
The network architecture of our 3D-RPN is depicted in~Figure~\ref{fig:3d-rpn}.
To encode the location information of each pair of anchors, we pass the fused VGG-16 feature map
through a $3 \times 3$ convolution~\textbf{(b)},
a rectified linear nonlinearity~\textbf{(c)},
and two more $1 \times 1$ convolution~(\textbf{(e)} and~\textbf{(h)}) layers. 
The first conv layer~\textbf{(b)} consists of 256 convolution filters
with $1 \times 1$ stride and padding, resulting in a feature map of size $[256 \times H' \times W']$~\textbf{(d)}.  
The second conv layer~\textbf{(e)} has $8 \times k$ convolution filters with $1 \times 1$ stride and does not have padding.
It outputs a feature map of shape $[(8 \times k) \times H' \times W']$~\textbf{(f)} which encodes the location information
(8 coordinate values) of $[ k \times H' \times W']$ pairs of anchor boxes~\textbf{(g)}.
The third conv layer~\textbf{(h)} is the same as~\textbf{(e)}. The only difference is in the 
number of filters which is $2 \times k$ to encode the actionness score (i.e. probability of action or no-action)
~\textbf{(j)}
for each $k$ pairs of anchors.

As RPN is a fully convolutional neural network, 
classification and regression weights are learned directly from the convolution features, whereas in
the fully connected layers (\S~\ref{subsec:fc-layers}) we apply linear transformation layers
for classification and regression.
In our 3D-RPN, the convolution layer~\textbf{(e)} is considered as the regression layer, as it outputs the 8 regression offsets
per pair of anchor boxes; the convolution layer~\textbf{(h)} is the classification layer.



\subsection{3D region proposal sampling} \label{subsec:prop-sampler}

Processing all the resulting region proposals is very expensive.
For example, with $k=12$ and a feature map of size $[512 \times 38 \times 50]$,
we get $12 \times 38 \times 50 = 22800$ pairs of anchor boxes.
For this reason, we subsample them during both training and testing
following the approach of~\cite{ren2015faster}~(\S~Figure~\ref{fig:algorithmOverview}~\textbf{(e)}).
We only make a slight modification in the sampling technique, as in our case one sample consists of a pair of bounding boxes, rather than a single box.

\textbf{Training time sampling.} During training, we compute the intersection over union (IoU) between
a pair of ground-truth boxes $\{G_{t}, G_{t+\Delta}\}$ and a pair of
proposal boxes $\{P_{1}, P_{2}\}$, so that,
$\psi_{1} = IoU(G_{t},P_{1})$ and $\psi_{2} = IoU(G_{t+\Delta},P_{2})$.
We consider $\{P_{1}, P_{2}\}$ as a positive example if $\psi_{1}>=0.5$ and  $\psi_{2}>=0.5$, that is 
both IoU values are above $0.5$.
When enforcing this condition, there might be cases in which we do not have any positive pairs.
To avoid such cases, we also consider as positive pairs those which have maximal mean
IoU $(\psi_{1}+\psi_{2})/2$ with the ground-truth pair. 
As negative examples we consider pairs for which both IoU values are below 0.3.

We construct a minibatch of size $B$  
in which we can have at most $B_{p} = B/2$ positive and 
$B_{N} = B - B_{P}$ negative training samples.
We set $B=256$.
Note that the ground-truth boxes $\{G_{t}, G_{t+\Delta}\}$ in a pair belong
to a same action instance but come from two different video frames $\{f_{t}, f_{t+\Delta}\}$.
As there may be multiple action instances present,
during sampling one needs to make sure that a pair of ground-truth boxes belongs to the same instance. 
To this purpose, we use the ground-truth tube-id provided in the datasets to keep track of instances.

\textbf{Test time sampling.} During testing, we use non-maximum suppression (NMS) to select the top $B = 1000$ proposal pairs.
We made changes to the NMS algorithm to select the top $B$ pairs of boxes based on their confidence.
In NMS, one first selects the box with the highest confidence, to then compute the IoU between
the selected box and the rest. 
In our modified version (i) we first select the pair of detection boxes 
with the highest confidence; (ii) we then compute the mean IoU between the selected pair and the remaining pairs,
and finally (iii) remove from the detection list pairs whose IoU is above an overlap threshold $th_{nms}$.

\subsection{Bilinear Interpolation} \label{bilinear-interpolation}

The sampled 3D region proposals are of different sizes and aspect ratios. 
We use \emph{bilinear interpolation}~\cite{gregor2015draw,jaderberg2015spatial}
to provide a fixed-size feature representation for them, necessary
to pass the feature map of each 3D region proposal to the fully connected layer fc6 of VGG-16 (\S~Figure~\ref{fig:algorithmOverview}~\textbf{(j)}),
which indeed requires a fixed-size feature map as input.

Whereas recent action detection methods \cite{peng2016eccv,Saha2016} use max-pooling of region of interest (RoI) features which only backpropagates the gradients w.r.t. convolutional features, bilinear interpolation allows us to backpropagate gradients with respect to both (a) convolutional features and (b) 3D RoI coordinates. Further, whereas \cite{peng2016eccv,Saha2016} train appearance and motion streams independently, and
perform fusion at test time, our model requires one-time training, and feature fusion is done at training time. 

\textbf{Feature fusion of 3D region proposals.} 
As a 3D proposal consists of a pair of bounding boxes, 
we apply bilinear feature pooling independently to each bounding box in the pair. This yields
two fixed-size pooled feature maps of size $[D \times kh \times kw]$
for each 3D proposal.
We then apply element-wise sum fusion (\S~Figure~\ref{fig:algorithmOverview}~\textbf{(i)}) 
to these $2$ feature maps, producing an output feature map of size $[D \times kh \times kw]$. 
Each fused feature map encodes the appearance and motion information of (the portion of) an action instance
which may be present within the corresponding 3D region proposal.
In this work, we use $D=512$, $kh$ $=$ $kw$ $=$ $7$.

\subsection{Fully connected layers} \label{subsec:fc-layers}

Our network employs two fully connected layers FC6 and FC7~(Figure~\ref{fig:algorithmOverview}~\textbf{(j)}),
followed by an action classification layer and a micro-tube regression layer (Figure~\ref{fig:algorithmOverview}~\textbf{(k)}).
\\
The fused feature maps~(\S~Section~\ref{bilinear-interpolation}) for each 3D proposal
are flattened into a vector and passed through FC6 and FC7.
Both layers use rectified linear units and dropout regularisation~\cite{Johnson2016}. 
For each 3D region proposal, the FC7 layer outputs a 4096 dimension feature vector
which encodes the appearance and motion features associated with the pair of bounding boxes.
Finally, these 4096-dimensional feature vectors are passed to the classification and regression layers.
The latter output $[B \times C]$ softmax scores and
$[B \times C \times 8]$ bounding box regression offsets (\S~\ref{subsec:3d-rpn}), respectively,
for $B$ predicted micro-tubes and $C$ action classes.

\section{Network training}
\vskip -2mm
\subsection{Multi-task loss function} \label{subsec:loss-func}

As can be observed in Figures \ref{fig:algorithmOverview} and \ref{fig:3d-rpn}, our network contains
two distinct classification layers.
\\
The \emph{mid classification layer} (\S~Figure~\ref{fig:3d-rpn}~\textbf{(h)}) predicts the probability $p^m$ of a 3D proposal containing an action, $p^{m} = (p^{m}_{0}, p^{m}_{1})$ 
over two classes (action vs. no action). We denote the associated loss by $L_{cls}^{m}$.
The \emph{end classification layer} (\S~Figure~\ref{fig:algorithmOverview}~\textbf{(k)})
outputs a discrete probability distribution (per 3D proposal),
$p^{e} = (p^{e}_{0},..., p^{e}_{C})$, over $C+1$ action categories. 
We denote the associated loss as $L_{cls}^{e}$.
\\
In the same way, the network has a mid (Figure~\ref{fig:3d-rpn}~\textbf{(e)}) and an end (\S~Figure~\ref{fig:algorithmOverview}~\textbf{(k)}) regression layer -- the associated
losses are denoted by $L_{loc}^{m}$ and $L_{loc}^{e}$, respectively.
Both regression layers output a pair of bounding box offsets $\phi^{m}$ and $\phi^{e}$ (cfr. Eq. \ref{eq:transform}).
We adopt the parameterization of $\phi$ (\S~\ref{subsec:3d-rpn}) given in~\cite{girshick-2014}.

Now, each training 3D proposal is labelled with a ground-truth action class $c^{e}$ and
a ground-truth micro-tube~(\S~\ref{sec:intro}) regression target $g^{e}$. We can then use the multi-task loss~\cite{ren2015faster}:
\begin{multline} \label{eqn:loss-function}
L(p^{e}, c^{e}, \phi^{e}, g^{e}, p^{m}, c^{m}, \phi^{m}, g^{m}) = \\
\lambda^{e}_{cls} L^{e}_{cls}(p^{e},c^{e}) + \lambda^{e}_{loc} [c \geq 1]L^{e}_{loc}(\phi^{e},g^{e}) + \\
\lambda^{m}_{cls} L^{m}_{cls}(p^{m},c^{m}) + \lambda^{m}_{loc} [c = 1]L^{m}_{loc}(\phi^{m},g^{m})\\
\end{multline}
on each labelled  3D proposal to jointly train for
(i) action classification ($p^e$),
(ii) micro-tube regression ($\phi^e$),
(iii) actionness classification ($p^m$), and
(iv) 3D proposal regression ($\phi^m$). 
Here, $L^{e}_{cls}(p^{e},c^{e})$ and $L^{m}_{cls}(p^{m},c^{m})$ are the cross-entropy losses for the true classes $c^{e}$ and $c^{m}$ respectively,  
where $c^{m}$ is $1$ if the 3D proposal is positive
and $0$ if it is negative, 
and $c^{e} = \{1,...,C\}$.
\\
The second term $L^{e}_{loc}(\phi^{e},g^{e})$ is defined over an 8-dim tuple of ground-truth micro-tube regression target coordinates: 
$
g^{e} = \Big (\{g^{e}_{x_{1}},g^{e}_{y_{1}},g^{e}_{w_{1}},g^{e}_{h_{1}}\},\{g^{e}_{x_{2}},g^{e}_{y_{2}},g^{e}_{w_{2}},g^{e}_{h_{2}}\} \Big )
$
and the corresponding predicted micro-tube tuple:
$
\phi^{e} = \Big (\{\phi^{e}_{x_{1}},\phi^{e}_{y_{1}},\phi^{e}_{w_{1}},\phi^{e}_{h_{1}}\},\{\phi^{e}_{x_{2}},\phi^{e}_{y_{2}},\phi^{e}_{w_{2}},\phi^{e}_{h_{2}}\} \Big ).
$
The fourth term $L^{m}_{loc}(\phi^{m},g^{m})$ is similarly defined over a tuple  $g^{m}$ of ground-truth 3D proposal regression target coordinates 
and the associated predicted tuple $\phi^{m}$.
The Iverson bracket indicator function $[c \geq 1]$ in (\ref{eqn:loss-function}) returns $1$ when $c^{e}\geq1$ and $0$ otherwise; $[c=1]$
returns $1$ when $c^{m}=1$ and $0$ otherwise.

For both regression layers we use a smooth $L1$ loss in transformed coordinate space as suggested by~\cite{ren2015faster}.
The hyper-parameters  
$\lambda^{e}_{cls}$,
$\lambda^{e}_{loc}$,
$\lambda^{m}_{cls}$ and
$\lambda^{m}_{loc}$,
in~Eq.~\ref{eqn:loss-function} 
weigh the relative importance of the four loss terms. In the following we set to 1 all four hyper-parameters.

\subsection{Optimisation}

We follow the end-to-end training strategy of~\cite{Johnson2016} to
train the entire network in a single optimisation step.
We use stochastic gradient descent (SGD) to update the weights of the two VGG-16 convolutional networks, with a momentum of $0.9$.
To update the weights of other layers of the network, we use the Adam~\cite{kingma2014adam} optimiser, with parameter values
$\beta_{1}=0.9$, $\beta_{2}=0.99$ and a learning rate of $1 \times 10^{-6}$.
During the 1st training epoch, we freeze the weights of the convolution networks and update only the weights
of the rest of the network.
We start fine-tuning the layers of the two parallel CNNs after completion of 1st epoch.
The first four layers of both CNNs are not fine-tuned for sake of efficiency.
The VGG-16 pretrained ImageNet weights are used to initialise the convolutional nets.
The rest of the network's weights are initialised using a Gaussian with $\sigma=0.01$. 

\section{Action-tube generation} \label{sec:action-tube}
\begin{figure}[h!]
  \centering
  \includegraphics[scale=0.25]{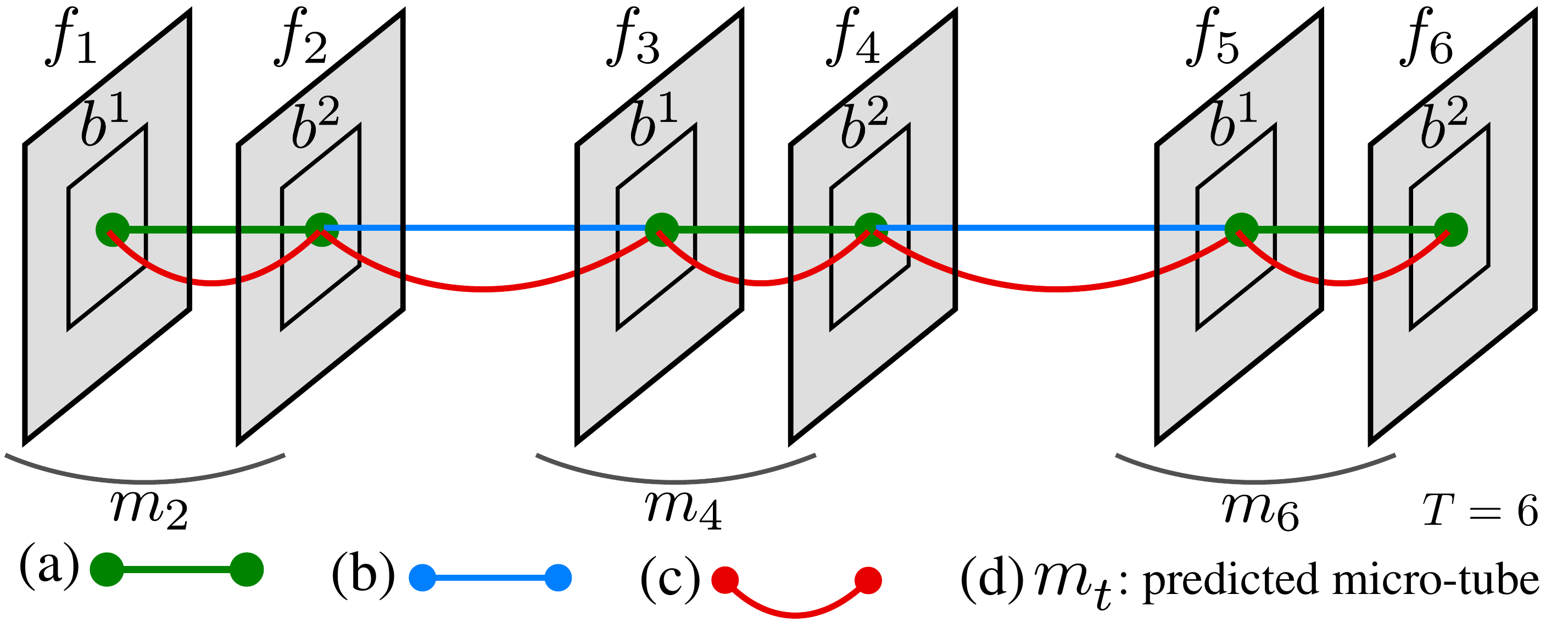}
  \vskip 2mm
  \caption{
  {\small
      \textit{
  \textbf{(a)} The temporal associations learned by our network;
  \textbf{(b)} Our micro-tube linking algorithm requires $(T/2 - 1)$ connections;
  \textbf{(c)} the $T-1$ connections required by \cite{Saha2016}'s approach.}
  }
  }  
  \label{fig:micro-tube-link} \vspace{-4mm}
\end{figure}
Once the predicted micro-tubes are regressed at test time, they need to be linked up to create complete action tubes associated with an action instance.
To do this we introduce here a new action tube generation algorithm which is an evolution of that presented in~\cite{Saha2016}. There, 
temporally untrimmed action paths are first generated
in a first pass of dynamic programming.  
In a second pass, paths are temporally trimmed to detect their start and end time. 
Here we modify the first pass of~\cite{Saha2016} and build action paths using the temporal associations learned by our network.
We use the second pass without any modification.

Linking up micro tubes (\S~Figure~\ref{fig:micro-tube-link}) is not the same as linking up frame-level detections as in~\cite{Saha2016}. 
In the Viterbi forward pass of \cite{Saha2016}, the edge scores between bounding boxes belonging to consecutive video frames (i.e., frame $f_{t}$ and $f_{t+1}$) are first computed. Subsequently, a DP (dynamic programming) matrix is constructed to keep track of the box indices with maximum edge scores.
In the Viterbi backward pass,
all consecutive pairs of frames, i.e, frames $\{1,2\}, \{2,3\}, \dots$ are traversed to join detections in time.
Our linking algorithm saves $50\%$ of the computing time,
by generating edge scores between micro-tubes (which only needs $T/2-1$ iterations, cfr. Figure \ref{fig:micro-tube-link}) rather than between boxes from consecutive frames (which, in the forward pass, needs $T-1$ iterations). In the backward pass, the algorithm connects the micro-tubes as per the max edge scores.

Recall that a predicted micro-tube consists of a pair of bounding boxes (\S~Figure~\ref{fig:micro-tube-link}), so that $m = \{b^{1},b^{2}\}$.
In the first pass action-specific paths $\Vector{p}_{c}=\{ {m}_{t}, t \in I = \{2,4,...,T-2\} \}$,
spanning the entire video length are obtained by maximising via dynamic programming~\cite{Georgia-2015a}:
\begin{equation}
  \label{eqn:firstpassenergy}
  E(\Vector{p}_c) = \sum_{t \in I} s_c({m}_t) + \lambda_{o} \sum_{t \in I} \psi_{o} \left({b^{2}}_{m_{t}}, {b^{1}}_{m_{t+2}}\right),
\end{equation}
where 
$s_c({m}_t)$ denotes the softmax score (\S~\ref{subsec:fc-layers})
of the predicted micro-tube $m$ at time step $t$,
the overlap potential $\psi_{o}({b^{2}}_{m_{t}}, {b^{1}}_{m_{t+2}})$ 
is the IoU between the second detection box ${b^{2}}_{m_{t}}$ which forms micro-tube $m_{t}$ and the
first detection box ${b^{1}}_{m_{t+2}}$ of micro-tube $m_{t+2}$. 
Finally, $\lambda_{o}$ is a scalar parameter weighting the relative importance of the pairwise term.
By recursively removing the detection micro-tubes associated with the current optimal path and maximising
(\ref{eqn:firstpassenergy}) for the remaining micro-tubes we can
account for multiple co-occurring instances of the same action class.


\section{Experiments} \label{sec:experiments}

\subsection{Experimental setting} \label{sec:data_eval_samp}

\textbf{Datasets.} All the experiments are conducted using the following two widely used action detection datasets: 
a) J-HMDB-21~\cite{J-HMDB-Jhuang-2013} and b) UCF-101 24-class 
\cite{soomro-2012}.

\textbf{J-HMDB-21} is a subset of the relatively larger action classification dataset HMDB-51~\cite{HMDBkuehne2011hmdb}, and is specifically designed for spatial action detection.
It consists of 928 video sequences and 21 different action categories.
All video sequences are temporally trimmed as per the action's duration,
and each sequence contains only one action instance.
Video duration varies from 15 to 40 frames. 
Ground-truth bounding boxes for human silhouettes are provided for all 21 classes, and the dataset is divided into 3 train and test splits.
For evaluation on J-HMDB-21 we average our results over the 3 splits.

The \textbf{UCF-101} 24-class action detection dataset is a subset of the larger UCF-101 action classification dataset, and comprises 24 action categories and 3207 videos for which spatiotemporal ground-truth annotations are provided. We conduct all our experiments using the first split. 
Compared to J-HMDB-21, the UCF-101 videos are relatively longer and temporally untrimmed, i.e., 
action detection is to be performed in both space and time.
Video duration ranges between 100 and 1000 video frames.

Note that the THUMOS~\cite{gorban2015thumos} and ActivityNet~\cite{caba2015activitynet} datasets are not suitable for spatiotemporal localisation, as they lack bounding box annotation.

\textbf{Evaluation metrics.}
As evaluation metrics we use both: 
(1) \emph{frame-AP} (the average precision of detections at the frame level) as in~\cite{Georgia-2015a,peng2016eccv};
(2) \emph{video-AP} (the average precision of detection at video level) as in~\cite{Georgia-2015a,Weinzaepfel-2015,Saha2016,peng2016eccv}. 
We select an IoU threshold ($\delta$) range [0.1:0.1:0.5] for J-HMDB-21 and [0.1,0.2,0.3] for UCF-101 when computing video-mAP. For frame-mAP evaluation we set $\delta =  0.5$. 

\textbf{Training data sampling strategy}.
As the input to our model is a pair of successive video frames and their associated ground-truth
{micro-tubes}, 
training data needs to be passed in a different way than in the
frame-level training approach~\cite{Georgia-2015a,Weinzaepfel-2015,peng2016eccv,Saha2016},
where inputs are individual video frames.
In our experiments, we use 3 different 
sampling schemes to construct training examples
using different combinations of successive video frames~(\S~Figure~\ref{fig:3d-prop}~\textbf{(b)}):
(1) \emph{scheme-11} 
generates training examples from the pairs of frames \{t=1,t=2\}, \{t=2,t=3\} \dots;
\emph{scheme-21} uses the (non-overlapping) pairs \{1,2\}, \{3,4\} \dots;
\emph{scheme-32} constructs training samples from the pairs \{1,3\}, \{4,6\} \dots

\subsection{Model evaluation} \label{subsec:model_eval}

We first show how a proper positive IoU threshold is essential during the
sampling of 3D region proposals at training time~(\S~\ref{subsec:prop-sampler}). 
Secondly, we assess whether our proposed network architecture,
coupled with the new data sampling strategies (Sec.~\ref{sec:data_eval_samp}), improves detection performance. 
We then show that our model outperforms the appearance-based model of~\cite{Saha2016}.
Finally, we compare the performance of the overall detection framework with the state-of-the-art.

\textbf{Effect of different positive IoU thresholds on detection performance.} 
We train our model on UCF-101 using two positive IoU thresholds: 0.7 and 0.5~(\S~\ref{subsec:prop-sampler}).
The detection results (video-mAP) of these two models ( Model-0-7 \& -0-5) are shown in Table~\ref{table:eff_diff_pos_iou}.  Whereas
\cite{ren2015faster} recommends an IoU threshold of 0.7 to subsample positive region proposals during training, 
in our case we observe that an IoU threshold of 0.5 works better with our model.
Indeed, during sampling we compute IoUs between pairs of bounding boxes and then take the mean IoU to subsample~(\S~\ref{subsec:prop-sampler}).
As the ground-truth boxes (micro-tubes) are connected in time and span different frames,
it is harder to get enough positive examples with a higher threshold like 0.7.
Therefore, in the remainder we use an IoU of 0.5 for evaluation.
\begin{table}[t] 
\centering
\caption{Effect of different positive IoU thresholds on detection performance (video-mAP).} 
\footnotesize
\vskip 2mm
\scalebox{0.85}{
\begin{tabular}{lccccc}
\toprule
 IoU threshold $\delta$  & 0.1              &  0.2             & 0.3               \\ \midrule  
 Model-0-7               & 64.04            & 54.83            & 44.664            \\
 Model-0-5               & \textbf{68.85}   & \textbf{60.06}   & \textbf{49.78}    \\
 \bottomrule
 \multicolumn{4}{l}{} 
\end{tabular}
}
\label{table:eff_diff_pos_iou}
\vspace{-5mm}
\end{table} 

\textbf{Effect of our training data sampling strategy on detection performance.} 
\emph{JHMDB-21 frame-mAP.}
We first generate a J-HMDB-21 training set 
using the \emph{scheme-11}~(\S~\ref{sec:data_eval_samp}) and train our model.
We then generate another training set 
using \emph{scheme-32}, 
and train our model on the combined training set (\emph{set-11+32}). 
Table~\ref{table:eff_diff_pair_comb_jhmdb_frameAP} shows the per class frame-AP
obtained using these two models.
We can observe that out of 21 JHMDB action classes, the frame-APs of 15 classes actually improve when training the model on the new combined trainset (\emph{set-11+32}).
Overall performance increases by 1.64\%, 
indicating that the network learns temporal association more efficiently when
it is trained on pairs generated from different combinations of successive video frames.

\emph{JHMDB-21 video-mAP.}
The two above trained models are denoted by \emph{Model-11} and \emph{Model-11+32} in Table~\ref{table:eff_diff_pair_comb}, where
the video-mAPs at different IoU threshold for these two models are shown.
Although the first training strategy \emph{scheme-11} already makes use of all the video frames present in J-HMDB-21 training splits,
when training our model using the combined trainset
we observe an improvement in the video-mAP of $1.04\%$ at $\delta=0.5$. 

\begin{table*}[!htbp] 
\centering
\caption{Effect of our training data sampling strategy on per class frame-AP at IoU threshold $\delta=0.5$, JHMDB-21 (averaged over 3 splits).} 
\footnotesize
\vskip 2mm
\scalebox{0.65}{
\begin{tabular}{l|ccccccccccccccccccccc|c}
\toprule
 \textbf{frame-AP(\%)} & brushHair & catch & clap & climbStairs & golf & jump & kickBall & pick & pour & pullup & push & run & shootBall & shootBow & shootGun & sit & stand & swingBaseball & throw & walk & wave & mAP	                \\ \midrule 
 
 ours (*) & 46.4  & 40.7  & 31.9  & 62.3  & 91.0  & 4.3  & 17.3  & 29.5  & 86.2  & 82.7  & 66.9  & 35.5  & 33.9  & 78.2 & 49.7  & 11.7  & 13.8  & 57.1  & 21.3  & 27.8  & 27.1 & 43.6 \\
 ours (**) & 43.7  & 43.6  & 33.0  & 61.5  & 91.8  & 5.6  & 23.8  & 31.5  & 91.8  & 84.1  & 73.1  & 32.3  & 33.3  & 81.4  & 55.1  & 12.4  & 14.7  & 56.3  & 22.2  & 24.7  & 29.4 & 45.0 \\ 
Improvement & -2.6  & \textbf{2.9}  & \textbf{1.0}  & -0.8  & \textbf{0.7}  & \textbf{1.2}  & \textbf{6.4}  & \textbf{1.9}  & \textbf{5.5}  & \textbf{1.4}  & \textbf{6.1}  & -3.2  & -0.6  & \textbf{3.2}  & \textbf{5.4}  & \textbf{0.6}  & \textbf{0.8}  & -0.8  & \textbf{0.8}  & -3.1  & \textbf{2.3} & \textbf{1.4} \\ \midrule
\cite{Georgia-2015a} & 65.2 & 18.3 & 38.1 & 39.0 & 79.4 & 7.3 & 9.4 & 25.2 & 80.2 & 82.8 & 33.6 & 11.6 & 5.6 & 66.8 & 27.0 & 32.1 & 34.2 & 33.6 & 15.5 & 34.0 & 21.9 & 36.2 \\
\cite{wangcvpr2016} & 60.1 & 34.2 & 56.4 & 38.9 & 83.1 & 10.8 & 24.5 & 38.5 & 71.5 & 67.5 & 21.3 & 19.8 & 11.6 & 78.0 & 50.6 & 10.9 & 43.0 & 48.9 & 26.5 & 25.2 & 15.8 & 39.9 \\
\cite{Weinzaepfel-2015} & 73.3 & 34.0 & 40.8 & 56.8 & 93.9 & 5.9 & 13.8 & 38.5 & 88.1 & 89.4 & 60.5 & 21.1 & 23.9 & 85.6 & 37.8 & 34.9 & 49.2 & 36.7 & 16.8 & 40.5 & 20.5 & 45.8 \\ 
\cite{peng2016eccv} & 75.8 & 38.4 & 62.2 & 62.4 & 99.6 & 12.7 & 35.1 & 57.8 & 96.8 & 97.3 & 79.6 & 38.1 & 52.8 & 90.8 & 62.7 & 33.6 & 48.9 & 62.2 & 25.6 & 59.7 & 37.1 & 58.5 \\ \midrule \midrule 
\textbf{video-AP(\%)} &  &  &  &  &  &  &  &  &  &  &  &  &  &  &  &  &  &  &  &  &  &  \\ \midrule 
ours (*) & 53.9  & 54.4  & 39.8  & 68.2  & 96.1  & 5.69  & 39.6  & 34.9  & 97.1  & 93.5  & 84.1  & 53.7  & 43.6  & 93.2  & 64.5  & 20.9  & 22.8  & 72.1  & 23.2  & 39.4  & 37.8 & 54.27 \\ 
ours (**) & 51.9  & 54.5  & 41.2  & 66.6  & 94.8  & 7.8 & 48.7  & 33.7  & 97.6  & 92.5  & 87.6  & 49.0  & 37.4  & 92.7  & 75.8  & 21.6  & 27.1  & 73.3 & 24.3  & 37.7  & 44.7 & 55.31 \\
Improvement & -1.9  & \textbf{0.01}  & \textbf{1.4}  & -1.6  & -1.2  & \textbf{2.1}  & \textbf{9.1}  & -1.2  & \textbf{0.4}  & -1.0  & \textbf{3.4}  & -4.7  & -6.2  & -0.5  & \textbf{11.2}  & \textbf{0.6}  & \textbf{4.2}  & \textbf{1.1}  & \textbf{1.1}  & -1.6  & \textbf{6.8} & \textbf{1.04} \\  \midrule 
\cite{Georgia-2015a} & 79.1 & 33.4 & 53.9 & 60.3 & 99.3 & 18.4 & 26.2 & 42.0 & 92.8 & 98.1 & 29.6 & 24.6 & 13.7 & 92.9 & 42.3 & 67.2 & 57.6 & 66.5 & 27.9 & 58.9 & 35.8 & 53.3 \\
\cite{wangcvpr2016} & 76.4 & 49.7 & 80.3 & 43.0 & 92.5 & 24.2 & 57.7 & 70.5 & 78.7 & 77.2 & 31.7 & 35.7 & 27.0 & 88.8 & 76.9 & 29.8 & 68.6 & 72.8 & 31.5 & 44.4 &  26.2 & 56.4 \\

\bottomrule
 \multicolumn{23}{l}{*Model-11 **Model-11+32} 
\end{tabular}
}
\label{table:eff_diff_pair_comb_jhmdb_frameAP}
\vspace{-3mm}
\end{table*} 

\begin{table*}[!htbp] 
\centering
\caption{Per class video-AP comparison at IoU threshold $\delta=0.2$, UCF-101.} 
\footnotesize
\vskip 2mm
\scalebox{0.65}{
\begin{tabular}{l|cccccccccccccc|c}
\toprule
\textbf{video-AP(\%)} & BasketballDunk & Biking & Diving & Fencing & FloorGymnastics & GolfSwing & IceDancing & LongJump & PoleVault & RopeClimbing & Skiing & Skijet & SoccerJuggling & WalkingWithDog & mAP\\ \midrule
\cite{Saha2016} (*A) & 22.7 & 56.1 & 89.7 & 86.9 & 93.8 & 59.9 & 59.2 & 41.5 & 48.9 & 77.8 & 68.4 & 88 & 34.6 & 73.3 & 56.86 \\
ours & 39.6 & 59.5 & 91.2 & 88.5 & 94.1 & 70.7 & 70.4 & 49.8 & 71.0 & 97.2 & 74.0 & 92.9 & 80.2 & 73.6 & 60.06 \\
Improvement & \textbf{16.9} & \textbf{3.4} & \textbf{1.5} & \textbf{1.6} & \textbf{0.3} & \textbf{10.8} & \textbf{11.2} & \textbf{8.3} & \textbf{22.1} & \textbf{19.4} & \textbf{5.6} & \textbf{4.9} & \textbf{45.6} & \textbf{0.3} & \textbf{3.2} \\
 \bottomrule
 \multicolumn{16}{l}{*A: appearance model} 
\end{tabular}
}
\label{table:efffect_app_feature_UCF101}
\vspace{-3mm}
\end{table*} 

\begin{table}[!htbp] 
\centering
\caption{Effect of our training data sampling strategy on video-mAP, JHMDB-21 (averaged over 3 splits).} 
\footnotesize
\vskip 2mm
\scalebox{0.8}{
\begin{tabular}{lccccc}
\toprule
 IoU threshold $\delta$ 	              & 0.1              &  0.2             & 0.3              & 0.4                     & 0.5  \\ \midrule 
 $\textnormal{Model-11}$              & 57.73            & 57.70            & 57.60            & \textbf{56.81}                   & 54.27   \\
 $\textnormal{Model-11+32}$           & \textbf{57.79}   & \textbf{57.76}   & \textbf{57.68}   & 56.79          & \textbf{55.31}   \\\midrule
 \bottomrule
 \multicolumn{6}{l}{} 
\end{tabular}
}
\label{table:eff_diff_pair_comb}
\vspace{-3mm}
\end{table} 

\begin{table}[!htbp] 
\centering
\caption{Spatio-temporal action detection performance (video-mAP) comparison with the state-of-the-art on J-HMDB-21.} 
\footnotesize
\vskip 2mm
\scalebox{0.8}{
\begin{tabular}{lccccc}
\toprule
 IoU threshold $\delta$ 			& 0.1    & 0.2    & 0.3   & 0.4   & 0.5  \\ \midrule
 Gkioxari and Malik~\cite{Georgia-2015a} 	& --     & --     & --    & --    & 53.30   \\
 Wang~\etal~\cite{wangcvpr2016}   		& --     & --     & --    & --    & 56.40   \\
 Weinzaepfel~\etal~\cite{Weinzaepfel-2015}	& --     & 63.1   & --    &  --   & 60.70   \\
 Saha ~\etal~\cite{Saha2016} (Spatial Model)	& 52.99  & 52.94  & 52.57 & 52.22 & 51.34   \\
 Peng and Schmid~\cite{peng2016eccv}		& --     &\textbf{74.3} & -- & -- & \textbf{73.1}\\\midrule
 $\textnormal{Ours}$                           & 57.79   & 57.76   & 57.68  & 56.79   & 55.31    \\\bottomrule 
\multicolumn{6}{l}{} 
\end{tabular}
}
\label{table:jhmdb21_results2}
\vspace{-3mm}
\end{table} 

\begin{table}[!htbp] 
  \centering
  \caption{Spatio-temporal action detection performance (video-mAP) comparison with the state-of-the-art on UCF-101.}
  \vskip 2mm
  {\footnotesize
  \scalebox{0.8}{
  \begin{tabular}{lcccccc}
  \toprule
  IoU threshold $\delta$ 			 & 0.1    & 0.2    & 0.3 & 0.5 & 0.75 & 0.5:0.95 \\ \midrule
  Yu \etal~\cite{yu2015fast} 			 & 42.8   & 26.50  & 14.6 & -- & -- & --\\
  Weinzaepfel~\etal~\cite{Weinzaepfel-2015} 	 & 51.7   & 46.8  &  37.8 & -- & -- & --\\   
  Peng and Schmid~\cite{peng2016eccv} 		 & \textbf{77.31}  & \textbf{72.86}  & \textbf{65.70} & 30.87 & \textbf{01.01} & 07.11\\
  Saha ~\etal~\cite{Saha2016} (*A)         	 & 65.45  & 56.55  & 48.52 & -- & -- & --\\
  Saha ~\etal~\cite{Saha2016} (full)         	 & 76.12  & 66.36  & 54.93 & -- & -- & --\\
  $\textnormal{Ours}-ML$                            & 68.85  & 60.06  & 49.78 & -- & -- & --\\
  $\textnormal{Ours}-ML-(*)$                     & 70.71  & 61.36  & 50.44 &  32.01 & 0.4 & 9.68\\
  $\textnormal{Ours}-2PDP-(*)$                   & 71.3   & 63.06  & 51.57 & \textbf{33.06} & 0.52 & \textbf{10.72} \\
    \bottomrule 
\multicolumn{7}{l}{(*) cross validated alphas as in~\cite{Saha2016}; 2PDP - tube generation algorithm~\cite{Saha2016}} \\
\multicolumn{7}{l}{ML - our micro-tube linking algorithm.} 
  \end{tabular}
  }
  }
  \label{table:ucf101_results2} \vspace{-3mm}
\end{table} 

\textbf{Effect of exploiting appearance features.}
Further, we show that our model exploits appearance features (raw RGB frames) efficiently, 
contributing to an improvement of video-mAP by $3.2\%$ over~\cite{Saha2016}.
We generate a training set for UCF-101 split 1 using the training \emph{scheme-21}
and compare our model's performance with that of the appearance-based model (*A) of~\cite{Saha2016}.
We show the comparison in Table~\ref{table:efffect_app_feature_UCF101}.

Note that, among the 24 UCF-101 action classes, our model exhibits better video-APs for 14 classes, with an overall gain of 3.2\%.
We can observe that, although trained on appearance features only, our model improves the video-APs significantly for action classes which exhibit a 
large variability in appearance and motion. 
Also, our model achieves relatively better spatiotemporal detection on action classes associated with  video sequences which are significantly temporally untrimmed,
such as \emph{BasketballDunk}, \emph{GolfSwing}, \emph{Diving} with relative video-AP improvements of 16.9\%, 10.8\% and 1.5\% respectively. 
We report significant gains in absolute video-AP for action categories \emph{SoccerJuggling}, \emph{PoleVault}, \emph{RopeClimbing}, \emph{BasketballDunk},
\emph{IceDancing}, \emph{GolfSwing} and \emph{LongJump}
of 45.6\%, 22.1\%, 19.4\%, 16.9\%, 11.2\% 10.8\% and 8.3\%, respectively.

\textbf{Detection performance comparison with the state-of-the-art.} Table~\ref{table:jhmdb21_results2} reports action detection results,
averaged over the three splits of \textit{J-HMDB-21}, 
and compares them with those to our closest competitors.
Note that, although our model only trained using the appearance features (RGB images), it outperforms~\cite{Georgia-2015a} which was trained using both appearance and optical flow features. Also, our model outperforms~\cite{Saha2016}'s spatial detection network.

Table~\ref{table:ucf101_results2} compares the action detection performance of our model on the UCF-101 dataset
to that of current state of the art approaches.
We can observe that our model outperforms~\cite{yu2015fast, Weinzaepfel-2015, Saha2016} by a large margin. In particular,
our appearance-based model outperforms~\cite{Weinzaepfel-2015} which exploits both appearance and flow features.
Also notice, our method works better than that of~\cite{peng2016eccv} at higher IoU threshold, which is more useful in real-world applications.

\section{Implementation details} \label{sec:imp_details}
We implement our method using Torch 7~\cite{collobert2011torch7}.
To develop our codebase, we take coding reference from the publicly available repository~\cite{densecap-git-repo}.
We use the coding implementation of bilinear interpolation~\cite{biIntPolationGitGub} (\S~Section~\ref{bilinear-interpolation})
for ROI feature pooling. 
Our micro-tube linking algorithm (ML) (\S~Section~\ref{sec:action-tube}) is implemented in MATLAB.\\

In all our experiments, at training time we pick top 2000 RPN generated 3D proposals using NMS (non-maximum suppression).
At test time we select top 1000 3D proposals.
However, a lower number of proposals, e.g. top 300 proposals does not effect the detection performance,
and increase the test time detection speed significantly.
In Section~\ref{subsec:num_3d_prop}, we show that 
extracting less number of 3D proposals (at test time) does not effect the detection performance.
Shaoqing \etal~\cite{ren2015faster} observed the same with Faster-RCNN.\\

For UCF-101, we report test time detection results (video-mAP) using two different
action-tube generation algorithms.
Firstly, we link the micro-tubes predicted by the proposed model (at test time)
using our \textbf{m}icro-tube \textbf{l}inking (ML) algorithm (\S~Section~\ref{sec:action-tube}).
we denote this as \emph{``Ours-ML''} in Table~\ref{table:ucf101_results2}.
Secondly, we construct final action-tubes from the predicted micro-tubes using
the \textbf{2} \textbf{p}ass \textbf{d}ynamic  \textbf{p}rogramming (2PDP) 
algorithm proposed by~\cite{Saha2016}.
We denote this as \emph{``Ours-2PDP''} in Table~\ref{table:ucf101_results2}.
The results in Table~\ref{table:eff_diff_pos_iou}, \ref{table:efffect_app_feature_UCF101}, \ref{table:eff_diff_pair_comb} and
\ref{table:jhmdb21_results2} are generated using our new micro-tube linking algorithm (\emph{``Ours-ML''}).
Further, we cross-validate the class-specific $\alpha_{c}$ as in Section 3.4 of~\cite{Saha2016},
and generate action-tubes using these cross-validated $\alpha_{c}$ values.
We denote the respective results using an asterisk (\textbf{`*'}) symbol in Table~\ref{table:ucf101_results2}.

\subsection{Mini-batch sampling}
In a similar fashion~\cite{girshick2015fast},
we construct our gradient descent mini-batches
by first sampling $N$ pairs of successive video frames, and then sampling
$R$ 3D proposals for each pair.
In practice, we set $N=1$ and $R=256$ in all our experiments.
We had one concern over this way of sampling training examples because,
all the positive 3D proposals from a single training batch (i.e. a pair of video frames)
belong to only one action category
\footnote{Each video clip of UCF-101 and J-HMDB-21 is associated with a single class label.
Therefore, a pair of video frames belongs to a single action class.}
(that is, they are correlated),
which may cause slow training convergence.
However, we experience a fast training convergence and good detection results with the above sampling strategy.
\subsection{Data preprocessing}
The dimension of each video frame in both J-HMDB-21 and UCF-101 is $[320 \times 240]$.
We scale up each frame to dimension $[800 \times 600]$ as in~\cite{ren2015faster}.
Then we swap the RGB channels to BGR and subtract the VGG image mean 
$\{103.939, 116.779, 123.68\}$ from each BGR pixel value.
\subsection{Data augmentation}
We augment the training sets by flipping each video frame horizontally with a probability of $0.5$.
\subsection{Training batch} Our training data loader script constructs a training batch which consists of:
a) a tensor of size $[2 \times D \times H \times W]$ containing the raw RGB pixel data for a pair of video frames,
where $D = 3$ refers to the 3 channel RGB data, $H = 600$ is the image height and $W = 800$ is the image width;
b) a tensor of size $[2 \times T \times 6]$ which contains the ground-truth {micro-tube}
annotation in the following format:
$[$\emph{fno} \emph{tid} $x_{c}$ $y_{c}$ $w$ $h]$, where $T$ is the number of {micro-tubes},
\emph{fno} is the frame number of the video frame,
\emph{tid} is an unique identification number assigned to each individual action tube
within a video, $\{x_{c},y_{c}\}$ is the center and
$w$ and $h$ are the width and height of the ground-truth bounding box;
c) a $[1 \times T]$ tensor storing the action class label
for each {micro-tube}. 
The J-HMDB-21 (Model-11+32) train set has 58k training batches,
and UCF-101 train set consists of 340k training batches.

\subsection{Training iteration}
Our model requires at least 2 training epochs
because,
in the first training epoch
we freeze the weights of all the convolutional layers and only 
update the weights of the rest of the network.
We start updating the weights of the convolutional layers (alongside other layers) in the second epoch.
We stop the training after 195k and 840k iterations for J-HMDB-21 and UCF-101 respectively.
The training times required for J-HMDB-21 and UCF-101 are 36 and 96 GPU hours respectively using a single GPU.
The training time can be further reduced by using two or more GPUs in parallel.
\section{Fusion methods}
A fusion function $f$ : $\vec{x}^{t}$,  $\vec{x}^{t+\Delta}$, $\rightarrow y$ fuses two convolution feature maps
$\vec{x}^{t}, \vec{x}^{t+\Delta} \in \mathbb{R}^{H' \times W' \times D}$ to produce an output 
map $y \in \mathbb{R}^{H' \times W' \times D}$, where $W'$, $H'$ and $D$ are the width, height and number of
channels of the respective feature maps~\cite{Feichtenhofer-2016}. In this work we experiment with 
the following two fusion methods.\\

\textbf{Sum fusion.}
Sum fusion $y^{sum} = f^{sum}(\vec{x}^{t}, \vec{x}^{t+\Delta})$ computes the sum of the two feature maps
at the same spatial locations, $(i,j)$ and feature channels $d$:
\begin{align}
y^{sum}_{i,j,d} = \vec{x}^{t}_{i,j,d} + \vec{x}^{t+\Delta}_{i,j,d}
\end{align}
where $1 \leq i \leq H', 1 \leq j \leq W', 1 \leq d \leq D$ and
$\vec{x}^{t}, \vec{x}^{t+\Delta}, y \in \mathbb{R}^{H' \times W' \times D}$.\\

\textbf{Mean fusion.}
Mean fusion is same as sum fusion, only the difference is, instead of computing the element-wise sum, here we compute 
the element-wise mean:
\begin{align}
y^{mean}_{i,j,d} = (\vec{x}^{t}_{i,j,d} + \vec{x}^{t+\Delta}_{i,j,d}) / 2
\end{align}
\section{Additional experiments and discussion}
\subsection{Effect of different fusion methods}
In Table~\ref{table:eff_diff_fusion} we report video-mAPs obtained using mean and sum fusion methods for J-HMDB-21 dataset.
We train our model on the combined trainset (\emph{set-11+32}) (\S~Section \ref{sec:data_eval_samp} and \ref{subsec:model_eval}).
We train two models, one using mean and another using sum fusion and denote these two models in Table~\ref{table:eff_diff_fusion}
as \emph{Model-11+32 (mean-ML)} and \emph{Model-11+32 (sum-ML)} respectively.
Action-tubes are constructed using our micro-tube linking (ML) algorithm.
We can observe that at higher IoU threshold $\delta=0.5$, the sum fusion performs better and improve the mAP by almost $1\%$.
As a future work, we would like to explore different spatial and temporal feature map fusion functions~\cite{Feichtenhofer-2016}.
\begin{table}[h] 
\centering
\footnotesize
\caption{
{\small
 \textit{
Effect of element-wise mean and sum fusion methods on video-mAP for J-HMDB-21 dataset (averaged over 3 splits).
}}} 
\vskip 2mm
\scalebox{0.97}{
\begin{tabular}{lccccc}
\toprule
 IoU threshold $\delta$ 	      & 0.1              &  0.2             & 0.3              & 0.4                     & 0.5  \\ \midrule 
 \emph{Model-11+32 (mean-ML)}          & 57.16            & 57.14            & 57.00            & 56.13            & 54.51     \\
 \emph{Model-11+32 (sum-ML)}           & \textbf{57.79}   & \textbf{57.76}   & \textbf{57.68}   & \textbf{56.79}   & \textbf{55.31}   \\\midrule
 \bottomrule
 \multicolumn{6}{l}{} 
\end{tabular}
}
\label{table:eff_diff_fusion}
\vspace{-3mm}
\end{table} 
\subsection{Effect of the number of predicted 3D proposals} \label{subsec:num_3d_prop}
To investigate the effect of the number of predicted 3D proposals on detection performance,
we generate video-mAPs using two different sets of detections on J-HMDB-21 dataset. 
One detection set is generated by selecting top 1000 3D proposals 
and another set is by selecting top 300 3D proposals at test time using NMS.
Once the two sets of detections are extracted, predicted micro-tubes are then linked up in time to generate final action tubes.
Subsequently, video-mAPs are computed for each set of action tubes.
The corresponding video-mAPs for each detection set at different IoU thresholds are reported in Table~\ref{table:diff_num_prop}.
We denote these two detection sets in Table~\ref{table:diff_num_prop} as
\emph{Detection-1000}  and \emph{Detection-300}.
It is quite apparent that reduced number of RPN proposals does not effect the detection performance.
\begin{table}[h] 
\centering
\footnotesize
\caption{
{\small
 \textit{
Effect of the number of predicted 3D proposals on video-mAP for J-HMDB-21 dataset (averaged over 3 splits).
}}} 
\vskip 2mm
\scalebox{0.97}{
\begin{tabular}{lccccc}
\toprule
 IoU threshold $\delta$ 	            & 0.1                       &  0.2                       & 0.3                       & 0.4                      & 0.5  \\ \midrule  
 \emph{Detection-1000}           & 57.79                     & 57.76                     & 57.68                     & 56.79                     & \textbf{55.31}   \\
 \emph{Detection-300}           & \textbf{57.91}            & \textbf{57.89}            & \textbf{57.84}            & \textbf{56.87}            & 55.26  \\ \midrule
 \bottomrule
 \multicolumn{6}{l}{} 
\end{tabular}
}
\label{table:diff_num_prop}
\vspace{-3mm}
\end{table} 
\subsection{Loss function hyper-parameters} \label{subsec:loss_func_hyper_param}
We have four  hyper-parameters  
$\lambda^{e}_{cls}$,
$\lambda^{e}_{loc}$,
$\lambda^{m}_{cls}$ and
$\lambda^{m}_{loc}$,
in our multi-task loss function (\S~Equation~\ref{eqn:loss-function}) which
weigh the relative importance of the four loss terms.
To investigate the effect of these hyper-parameters on video-mAP,
we train our model with different combinations of these four hyper-parameters on J-HMDB-21 split-1.
The trainset is generated as per \emph{scheme-11} (\S~Section~\ref{sec:data_eval_samp}).
The video-mAPs of these trained models are presented in Table~\ref{table:loss_func_hp}.
We can observe that when the weigths for the \emph{mid classifcation} ($\lambda^{m}_{cls}$) and \emph{regression} ($\lambda^{m}_{loc}$) layers' loss terms
are too low (e.g. 0.1 \& 0.05), the model has the worst detection performance.
When all weights are set to 1, then the model exhibits good detection performance.
However, we get the best video-mAPs with $\lambda^{e}_{cls}=1.0$,  $\lambda^{e}_{loc}=1.0$, $\lambda^{m}_{cls}=0.5$ and  $\lambda^{m}_{loc}=0.5$.
In all our experiments we set all 4 weights to 1. As a future work, we will explore the setting $[1.0$, $1.0$, $0.5$, $0.5]$.
\begin{table}[h] 
\centering
\footnotesize
\caption{
{\small
 \textit{
Effect of different combinations of hyper-parameters on video-mAP for J-HMDB-21 split-1 train set.
}}} 
\vskip 2mm
\scalebox{0.97}{
\begin{tabular}{cccc|ccccc}
\toprule
\multicolumn{4}{c|}{Hyper-parameters} & \multicolumn{5}{c}{IoU threshold $\delta$} \\ 
$\lambda^{e}_{cls}$ & 	$\lambda^{e}_{loc}$ &  $\lambda^{m}_{cls}$ &   $\lambda^{m}_{loc}$        & 0.1      &  0.2     & 0.3      & 0.4    & 0.5  \\ \midrule  
 1.0                  &   1.0                   & 0.1                  & 0.05                     & 55.03    &  55.03   & 54.63    & 53.17  & 50.33 \\
 1.0                  &   1.0                   & 0.1                  & 0.1                      & 55.62    &  55.62   & 55.47    & 54.47  & 50.51 \\
 1.0                  &   1.0                   & 0.5                  & 0.25                     & 56.3     &  56.3    & 55.91    & 54.76  & 52.30 \\
 1.0                  &   1.0                   & 0.5                  & 0.5                      & \tb{57.3}&\tb{57.13}&\tb{56.79}&\tb{55.82}&\tb{53.81} \\
 1.0                  &   1.0                   & 1.0                  & 1.0                      & 56.86    &  56.85   & 56.57    & 55.89  & 52.78  \\\midrule  
 \bottomrule  
 \multicolumn{9}{l}{\emph{Model-11-2PDP}} \\ 
\end{tabular}
}
\label{table:loss_func_hp}
\vspace{-3mm}
\end{table} 
\subsection{Ablation study} 
An ablation study of the proposed model is presented in Section~\ref{subsec:discussion}.
Besides, as a part of the ablation study, per class frame- and video-APs of J-HMDB-21 dataset are reported in Table~\ref{table:eff_diff_pair_comb_jhmdb_frameAP},
and per class video-APs of UCF-101 are presented in Table~\ref{table:efffect_app_feature_UCF101} in the main paper.
\subsection{Discussion} \label{subsec:discussion}
The paper is about action detection, where evaluation is by class-wise average precision(AP) rather than classification accuracy, a confusion matrix cannot be used.
Our model is not limited to learn from pairs of consecutive frames, but can learn from pairs at any arbitrary interval $\Delta$ (see Figure~\ref{fig:3d-prop} \textbf{(a)}).

To confute this point we conducted an \textbf{ablation study} of our model which is discussed below.
For consecutive frames, we trained our model on J-HMDB-21 (split-01) dataset by passing training pairs composed of identical frames, 
e.g. passing the video frame pair $(65,65)$ instead of $(65,66)$.
As you can see in Table~\ref{table:ablation}, video-mAP drops significantly by 8.13\% (at IoU threshold $\delta=0.5$)
which implies that the two streams do not output identical representations.

To double-check, we also extracted the two VGG-16 conv feature maps (see Figure \ref{fig:algorithmOverview} \textbf{(b)})
for each test frame pair ($(f_{t}$, $f_{t+1})$) of J-HMDB-21 and UCF-101 datasets.
For each pair of conv feature maps, we first flattened them into feature vectors, and then computed the normalised $L2$ distance between them.
For identical frames we found that the $L2$ distance is $0$ for both J-HMDB-21 and UCF-101 datasets.
Whereas, for consecutive frames it is quite high, in case of J-HMDB-21 the mean $L2$ distance is $0.67$;
for UCF-101 the mean $L2$ distance is $0.77$ which again implies that the two streams generate significantly different feature encoding even for pairs consist of consecutive video frames.
\begin{table}[h] 
\centering
\footnotesize
\caption{
{\small
 \textit{
An ablation study on J-HMDB-21 (split-01). Video-mAP is computed at IoU threshold $\delta=0.5$.
}}} 
\vskip 2mm
\scalebox{0.97}{
\begin{tabular}{lc}
\toprule
Model & video-mAP (\%) \\ \midrule
\textbf{Model-01} & 48.9 \\
\textbf{Model-02} & 52.7 \\
\textbf{Model-03} & \textbf{57.1} \\ \midrule
\multicolumn{2}{l}{Model-01: Training pairs with identical frames} \\ 
\multicolumn{2}{l}{Model-02: Training pairs with consecutive frames (model-11)} \\ 
\multicolumn{2}{l}{Model-03: Training pairs with mixture of consecutive and} \\ 
\multicolumn{2}{l}{successive frames (model-11+32)} \\ \midrule
\bottomrule
\end{tabular}
}
\label{table:ablation}
\vspace{-3mm}
\end{table} 
\subsection{Computing time required for training/testing} \label{subsec:time_req}
\textbf{Computing time required for training.}
Saha \etal reported~\cite{Saha2016-main_sup_papers} that 
the state-of-the-art~\cite{Georgia-2015a,Weinzaepfel-2015} action detection methods 
require at least $6+$ days to train all the components (including fine-tuning CNNs, CNN feature extraction, one vs rest SVMs)
of their detection pipeline for UCF-101 trainset (split-01).
In our case, we need to train the model once which requires 96 hours for UCF-101 and 36 hours for J-HMDB-21 to train.
The training and test time calculations are done considering a single NVIDIA Titan X GPU.
The computing time requirement for different detection methods are presented in Table~\ref{table:computing_time_req}.
Our model requires 2 days less training time as compared to~\cite{Georgia-2015a,Weinzaepfel-2015} on UCF-101 trainset.
\\

\textbf{Computing time required for testing.}
We compare video-level computing time required (during test time)
of our method with~\cite{Georgia-2015a,Weinzaepfel-2015,Saha2016}  on J-HMDB-21 dataset.
Note that our method takes the least computing time of 8.5 Sec./video as compared to~\cite{Georgia-2015a,Weinzaepfel-2015,Saha2016} (\S~Table~\ref{table:computing_time_req}).
\begin{table}[h] 
\centering
\footnotesize
\caption{
{\small
 \textit{
Computing time comparison for training and testing.
}}} 
\vskip 2mm
\scalebox{0.97}{
\begin{tabular}{lcc}
\toprule
Methods                 & days (*)      &  Sec/video (**) \\ \midrule
\cite{Georgia-2015a}    & 6+            &  113.52 \\
\cite{Weinzaepfel-2015} & 6+            &  52.23 \\
\cite{Saha2016}         & \textbf{3+}   &  10.89 \\
\textbf{ours}           & 4             & \textbf{8.5}\\ \midrule
 \multicolumn{3}{l}{(*) Training time on UCF-101 dataset.} \\
 \multicolumn{3}{l}{(**) Average detection time on J-HMDB-21.} \\ \midrule
 \bottomrule
\end{tabular}
}
\label{table:computing_time_req}
\vspace{-3mm}
\end{table} 
\subsection{Qualitative results}
\begin{figure*}[h]
  \centering
  \includegraphics[scale=0.41]{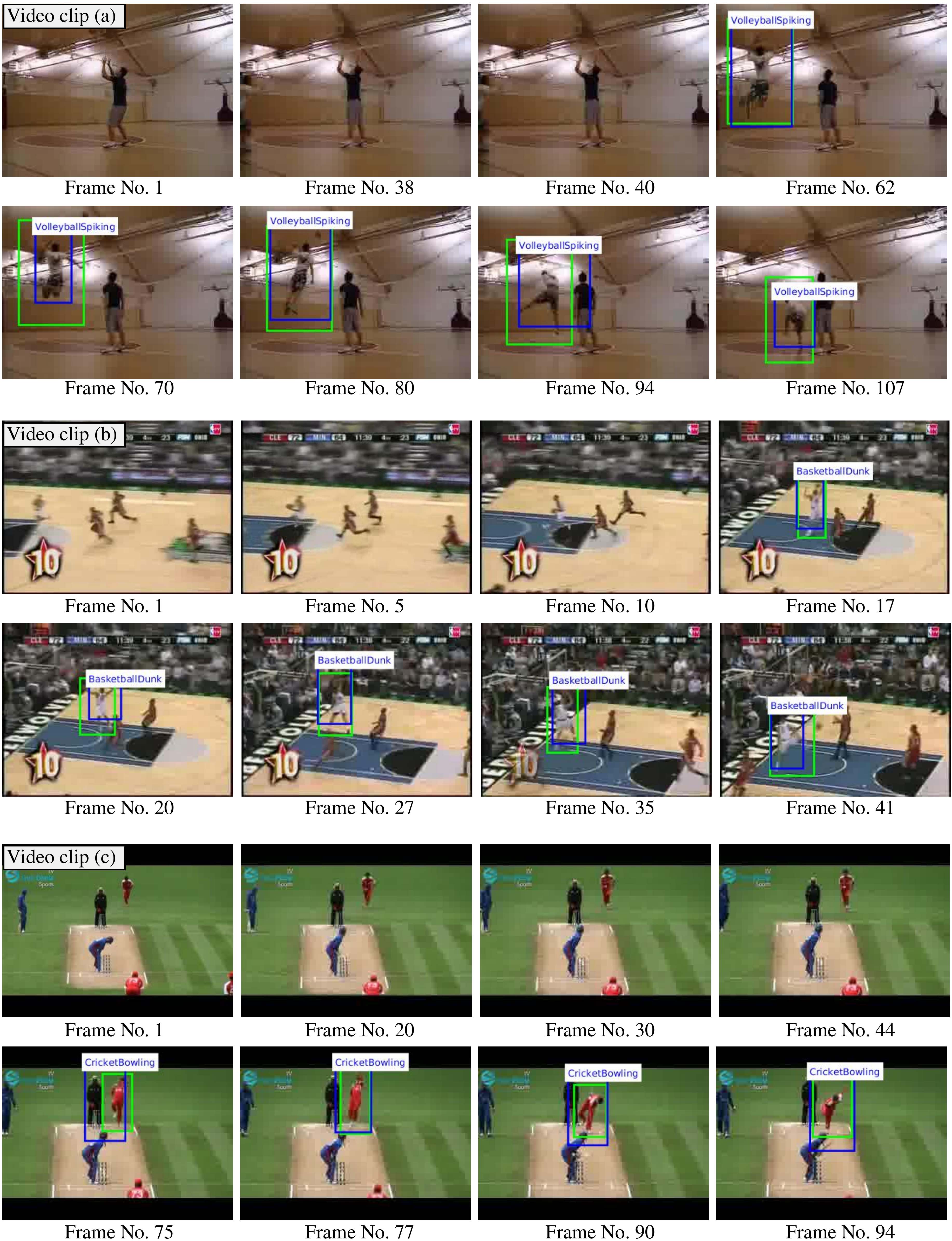}
  \vskip 2mm
  \caption{
   {\small
      \textit{
  Spatiotemporal action detection results.
  Video clips \textbf{(a)}, \textbf{(b)} and \textbf{(c)}
  are test videos belong to UCF-101 action classes \emph{VolleyballSpiking}, \emph{BasketballDunk} and \emph{CricketBowling} respectively.  
  }
  }
  }
  \label{fig:st-det-res-1} \vspace{-4mm}
\end{figure*}

\begin{figure*}[h]
  \centering
  \includegraphics[scale=0.41]{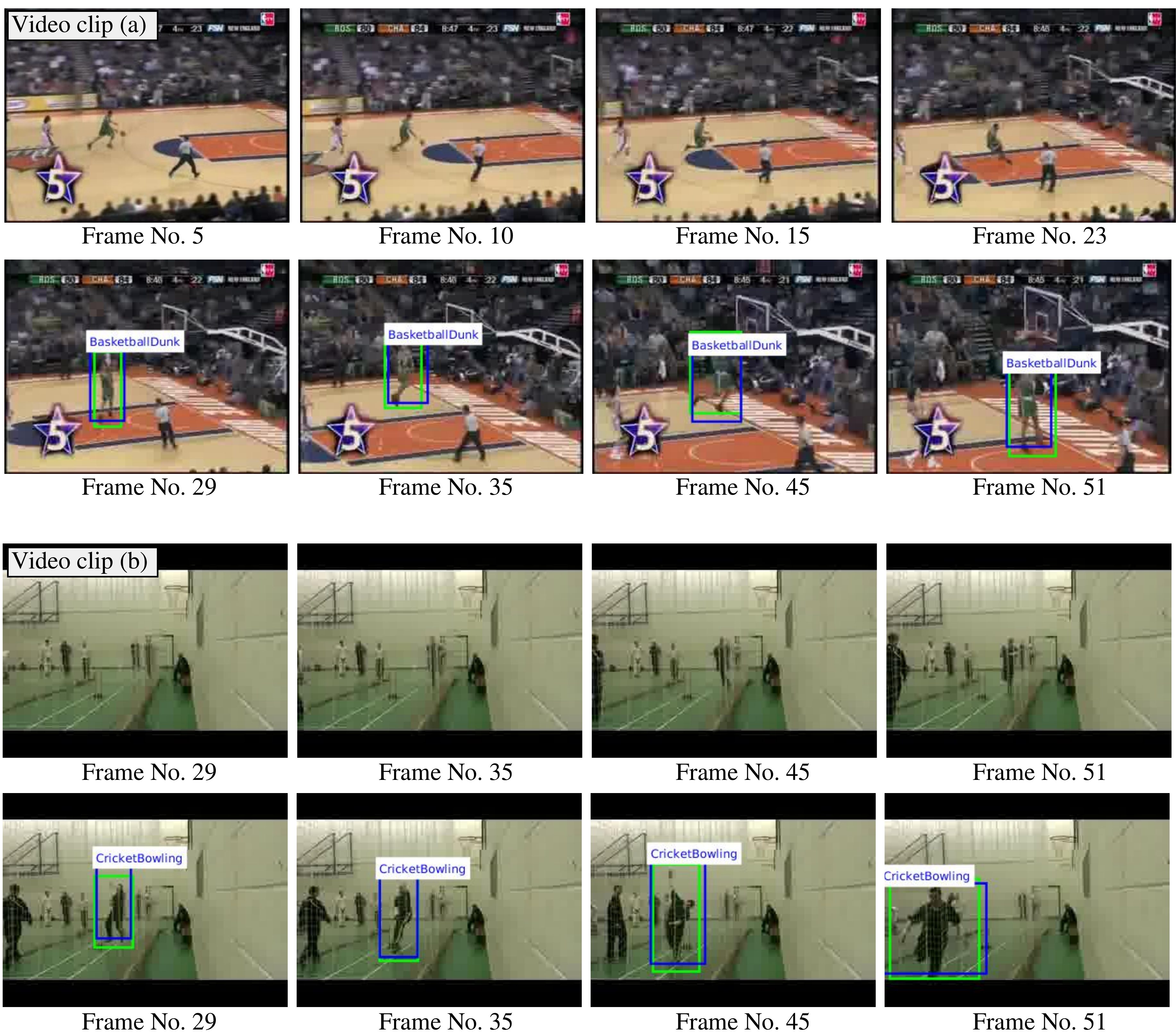}
  \vskip 2mm
  \caption{
   {\small
      \textit{
  Spatiotemporal action detection results. 
  Video clips \textbf{(a)} and \textbf{(b)}
  are test videos belong to UCF-101 action classes \emph{BasketballDunk} and \emph{CricketBowling} respectively.  
  }  
  }
  }
  \label{fig:st-det-res-2} \vspace{-4mm}
\end{figure*}

\textbf{Spatiotemporal action detection results on UCF-101. }
We show the spatiotemporal action detection qualitative results  
in Figures~\ref{fig:st-det-res-1} and ~\ref{fig:st-det-res-2}.
To demonstrate the robustness of the proposed detector against temporal action detection,
we select those action categories which have highly temporally untrimmed videos.
We select action classes \emph{VolleyballSpiking}, \emph{BasketballDunk} and \emph{CricketBowling}.
For \emph{VolleyballSpiking} class, 
the average temporal extent of the action in each video is 40\%,
that means, the remaining 60\% of the video doesn't contain any action.
Similarly, for \emph{BasketballDunk} and \emph{CricketBowling} classes,
we have average durations 41\% and 46\% respectively.

Video clip \textbf{(a)} (\S~Figures~\ref{fig:st-det-res-1})  has duration $107$ frames and the action \emph{VolleyballSpiking}
takes place only between frames $58$ to $107$. Note that our method able to successfully detect the temporal extent of the action 
(alongside spatial locations) which closely matches the ground-truth. We can observe similar quality of detection
results for video clip \textbf{(b)} and  \textbf{(c)} (\S~Figures~\ref{fig:st-det-res-1})
which have durations $41$ and $94$ frames and the temporal extent of
action instances are between frames $17$ to $41$ and frames $75$ to $94$ respectively for \emph{BasketballDunk} and \emph{CricketBowling}.
Video clips \textbf{(a)} and \textbf{(b)} in Figures~\ref{fig:st-det-res-2} show
some more spatiotemporal detection results for action classes \emph{BasketballDunk} and \emph{CricketBowling}.

\begin{figure*}[h]
  \centering
  \includegraphics[scale=0.41]{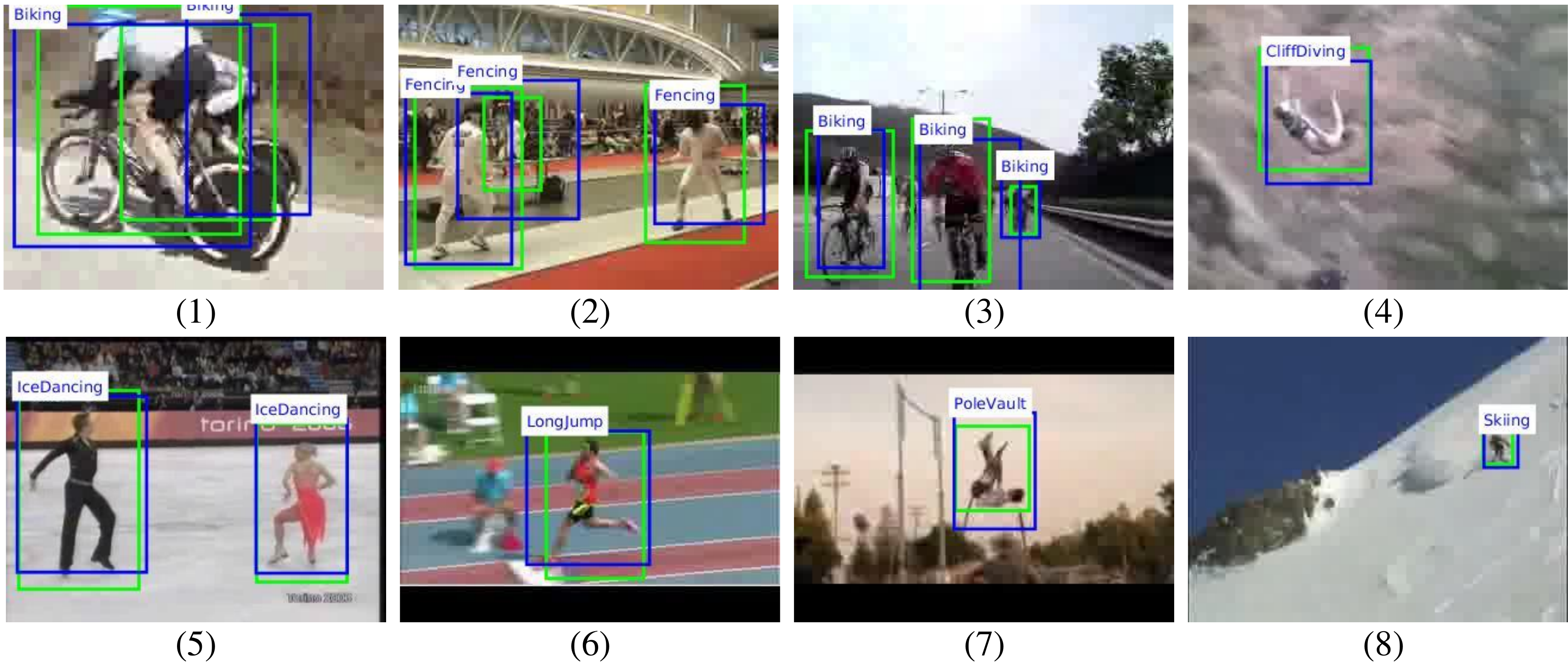}
  \vskip 2mm
  \caption{
   {\small
      \textit{
  More sample detection results on UCF-101 test videos.
  }
  }
  }
  \label{fig:st-det-res-3} \vspace{-4mm}
\end{figure*}

Figures~\ref{fig:st-det-res-3} shows sample detection results on UCF-101.
Note that in \textbf{(1)}, the 2nd ``biker'' is detected in spite of partial occlusion.
Figures~\ref{fig:st-det-res-3} \textbf{(1)}, \textbf{(2)}, \textbf{(3)} and \textbf{(5)} are examples
of multiple action instance detection with complex real world scenarios like 3 fencers (\S~\textbf{(2)}) and 
bikers (\S~\textbf{(3)}).
Further, note that the detector is robust against \emph{scale changes} as the 3rd fencer (\S~\textbf{(2)})
and the biker (\S~\textbf{(3)}) are detected accurately in spite of their relatively smaller shapes.\\

\begin{figure*}[h]
  \centering
  \includegraphics[scale=0.41]{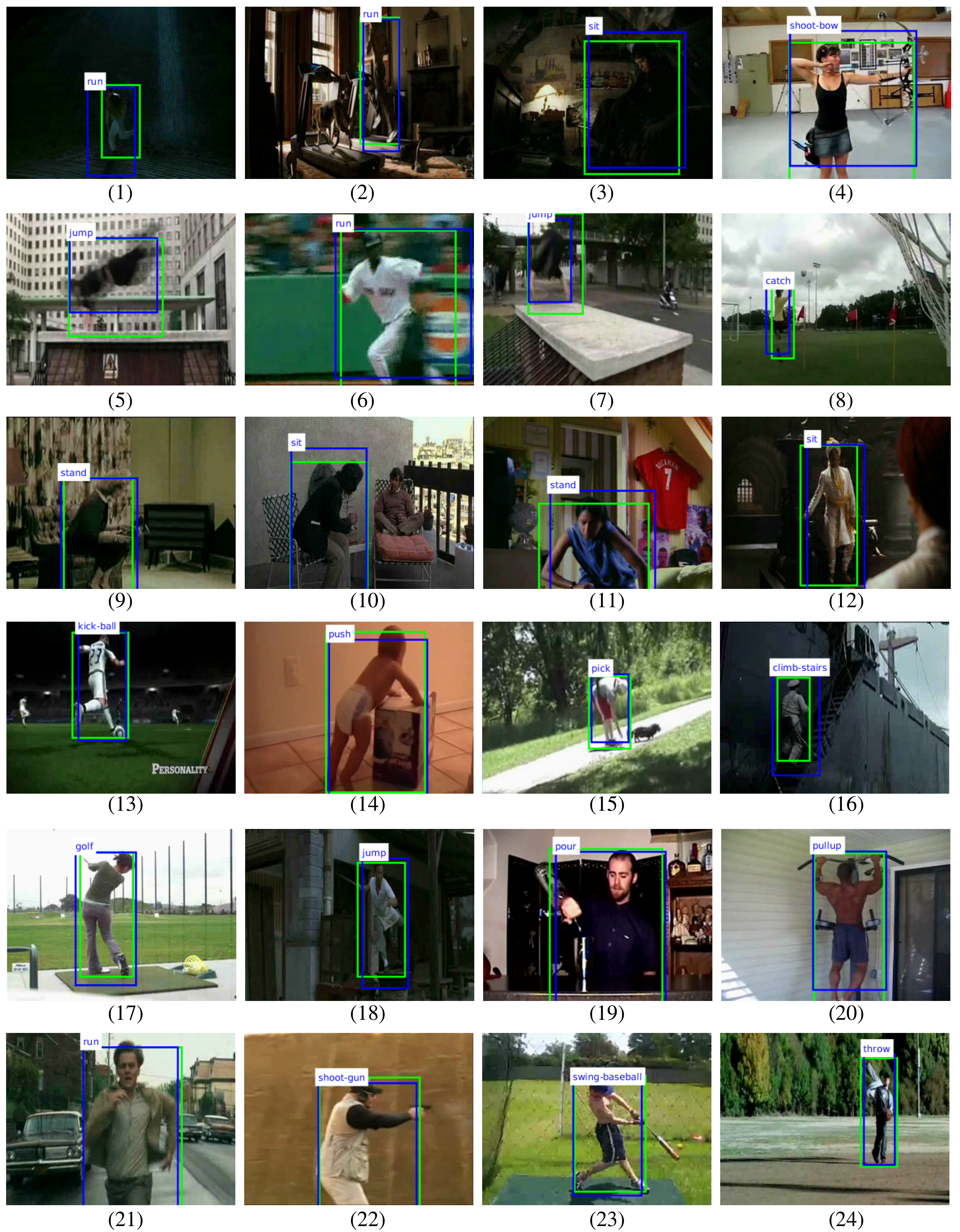}
  \vskip 2mm
  \caption{
   {\small
      \textit{
  Spatiotemporal action detection results on J-HMDB-21 test videos.
  }
  }
  }
  \label{fig:s-det-res-4} \vspace{-4mm}
\end{figure*}

\textbf{Spatiotemporal action detection results on J-HMDB-21. }
Figure~\ref{fig:s-det-res-4} presents the detection results of our model on J-HMDB-21 dataset.
In Figure~\ref{fig:s-det-res-4} \textbf{(1)}, \textbf{(2)} and \textbf{(3)}, the actions ``run'' and ``sit'' 
are detected accurately in spite of large variations in illumination conditions, which shows
that our detector is robust against \emph{illumination changes}.
In Figure~\ref{fig:s-det-res-4} \textbf{(5)}, \textbf{(6)} and \textbf{(7)}, the actions ``jump'' and ``run'' 
are detected successfully. Note that due to fast motion, these video frames are affected by \emph{motion blur}.
Further, in Figure~\ref{fig:s-det-res-4}  \textbf{(9)} to  \textbf{(12)}, actions ``stand'' and ``sit'' 
are detected with correct action labels. Even for human, it is hard to infer which instance belong to ``stand'' and 
``sit'' class. This again tells that our classifier is robust against inter-class similarity.
\section{Conclusions} \label{sec:conclusions}

In this work we departed from current practice in action detection to take a step towards deep network architectures able to classify and regress whole video subsets.
In particular, we propose a novel deep net framework able to regress and classify 3D region proposals spanning two successive video frames, effectively encoding the temporal aspect of actions using just raw RBG values. The proposed model is end-to-end trainable and can be jointly optimised for
action localisation and classification using a single step of optimisation. 
At test time the network predicts `micro-tubes' spanning two frames, which are linked up into complete action tubes via a new algorithm of our design.
Promising results confirm that
our model does indeed outperform the state-of-the-art when relying purely on appearance.

Much work will need to follow. It remains to be tested whether optical flow can be integrated in this framework and further boost performance. As the search space of 3D proposals is twice the dimension of that for 2D proposals, efficient parallelisation and search are crucial to fully exploit the potential of this approach. Further down the road we wish to extend the idea of micro-tubes to longer time intervals, posing severe challenges in terms of efficient regression in higher-dimensional spaces.
\vfill

\par\vfill\par
\clearpage
{\small
\bibliographystyle{ieee}
\bibliography{egbib}
}
\end{document}